\documentclass[12pt,reqno]{article}
\usepackage{arxiv}
\usepackage[letterpaper]{geometry}
\usepackage{amsmath,amssymb,amsfonts}
\usepackage{subcaption}
\usepackage{graphicx}
\usepackage{mathtools}

\graphicspath{{./}}

\usepackage{xcolor}

\newcommand{\fref}[1]{Fig.\,~\ref{#1}}
\newcommand{\tref}[1]{Table\,~\ref{#1}}
\newcommand{\eref}[1]{Eq.\,~(\ref{#1})}

\newcommand{\sref}[1]{Sec.\!~\ref{#1}}

\newcommand{\cref}[1]{Ref.\,~\cite{#1}}

\newcommand{\pdf}{{\pi}}

\newcommand{\cs}{\mathsf{c}}

\newcommand{\rs}{\mathsf{r}}

\newcommand{\gs}{\mathsf{g}}

\newcommand{\zs}{\mathsf{z}}
\newcommand{\As}{\mathsf{A}}

\newcommand{\Qs}{\mathsf{Q}}

\newcommand{\Rs}{\mathsf{R}}
\newcommand{\Hs}{\mathsf{H}}
\newcommand{\Is}{\mathsf{I}}

\newcommand{\Xs}{\mathsf{X}}
\newcommand{\Ks}{\mathsf{K}}
\newcommand{\Ls}{\mathsf{L}}
\newcommand{\Ps}{\mathsf{P}}

\newcommand{\eb}{\mathbf{e}}
\newcommand{\nb}{\mathbf{n}}

\newcommand{\xb}{\mathbf{x}}

\renewcommand{\sb}{\mathbf{s}}
\newcommand{\Ab}{\mathbf{A}}

\newcommand{\Lb}{\mathbf{L}}
\newcommand{\Rb}{\mathbf{R}}

\newcommand{\Sb}{\mathbf{S}}
\newcommand{\Ib}{\mathbf{I}}
\newcommand{\Xb}{\mathbf{X}}

\newcommand{\Ac}{\mathcal{A}}
\newcommand{\Uc}{\mathcal{U}}
\newcommand{\Nc}{\mathcal{N}}

\newcommand{\Cbb}{\mathbb{C}}

\newcommand{\varepsilonb}{\boldsymbol{\varepsilon}}

\newcommand{\Sigmab}{\boldsymbol{\Sigma}}

\newcommand{\phib}{{\boldsymbol{\phi}}}
\newcommand{\psib}{{\boldsymbol{\psi}}}
\newcommand{\Psib}{\boldsymbol{\Psi}}

\newcommand{\grad}{{\boldsymbol{\nabla}}}

\newcommand{\tr}{\operatorname{tr}}

\renewcommand{\div}{\operatorname{div}}
\newcommand{\divg}{\operatorname{div}_\mathcal{G}}
\newcommand{\Conv}{\operatorname{Conv}}

\newcommand{\dt}{\Delta\! t}

\newcommand{\momentum}{\mathsf{p}}
\newcommand{\NN}{\mathsf{N}\!\mathsf{N}}
\newcommand{\weight}{\mathsf{w}}
\newcommand{\data}{\mathcal{D}}
\newcommand{\inputvector}{\mathsf{X}}
\newcommand{\outputvector}{\mathsf{y}}
\newcommand{\outputvectorP}{\hat{\mathsf{y}}}
\newcommand{\hiddenvector}{\mathsf{h}}
\newcommand{\timevector}{\mathsf{f}}

\newcommand{\prob}{\pi}
\newcommand{\prior}{\pi}
\newcommand{\posterior}{\pi}

\newcommand{\giv}{\, | \,}

\newcommand{\hamiltonian}{H}
\newcommand{\potential}{\Phi}
\newcommand{\massmatrix}{\mathsf{M}}
\newcommand{\kernel}{k}

\newcommand{\strain}{\boldsymbol{\epsilon}}
\newcommand{\stress}{\mathbf{S}}
\newcommand{\EX}{\mathbb{E}}

\newcommand{\CDF}{\operatorname{CDF}}

\title{Uncertainty Quantification of Graph Convolution Neural Network Models of Evolving Processes}
\author{
Jeremiah Hauth  \\
University of Michigan,\\
Ann Arbor MI
\And
Cosmin Safta  \\
Sandia National Laboratories,\\
Livermore CA
\And
Xun Huan \\
University of Michigan, \\
Ann Arbor MI
\And
Ravi G. Patel \\
Sandia National Laboratories,\\ Albuquerque NM
\And
Reese E. Jones\thanks{corresponding: rjones@sandia.gov} \\
Sandia National Laboratories, \\
Livermore CA
}

\date{}

\renewcommand{\headeright}{}

\begin{document}

\maketitle

%\COMMENT{please provide ORCiD if you have one}

\begin{abstract}
The application of neural network models to scientific machine learning tasks has proliferated in recent years.
In particular, neural network models have proved to be adept at modeling processes with spatial-temporal complexity.
Nevertheless, these highly parameterized models have garnered skepticism in their ability to produce outputs with quantified error bounds over the regimes of interest.
Hence there is a need to find uncertainty quantification methods that are suitable for neural networks.
In this work we present comparisons of the parametric uncertainty quantification of neural networks modeling complex spatial-temporal processes with Hamiltonian Monte Carlo and Stein variational gradient descent and its projected variant.
Specifically we apply these methods to graph convolutional neural network models of evolving systems modelled with recurrent neural network and neural ordinary differential equations architectures.
We show that Stein variational inference is a viable alternative to Monte Carlo methods with some clear advantages for complex neural network models.
For our exemplars, Stein variational interference gave similar uncertainty profiles through time compared to Hamiltonian Monte Carlo, albeit with generally more generous variance.
Projected Stein variational gradient descent also produced similar uncertainty profiles to the non-projected counterpart, but large reductions in the active weight space were confounded by the stability of the neural network predictions and the convoluted likelihood landscape.
\end{abstract}

\section{Introduction}

Scientific machine learning (SciML)~\cite{Baker2019} seeks to combine principles of machine learning (ML) with scientific computing to tackle complex problems that are often multiscale and multiphysical, exposed to uncertainty and noise, and where decisions carry high consequences.
Neural networks (NNs) have emerged as a powerful tool in SciML due to their ability to learn and express complex nonlinear interactions.
NNs have been applied to a wide range of tasks in this context, such as in subgrid surrogate modeling~\cite{frankel2019oligocrystals,frankel2019evolution,frankel2022mesh,jones2023deep}, partial differential equations (PDEs) \cite{raissi2018deep,raissi2019physics,lu2019deeponet}, and reinforcement learning~\cite{banerjee2023survey,villarreal2023design,Shen2023}.
These NNs easily reach hundreds, and even millions of parameters~\cite{du2018gradient,soltanolkotabi2018theoretical,zou2019improved}.
While expressive, a highly parameterized NN also becomes prone to over-fitting and poor generalizability if not treated carefully.
Combined with the general lack of interpretability of NNs and the high-stakes nature of many SciML applications, a strong motivation emerges for uncertainty quantification (UQ) in NN models that is imperative for promoting  transparency of model limitations and strengthening the trust of their predictions.

We focus on Bayesian UQ~\cite{Berger1985,Sivia2006,vonToussaint2011} that seeks to capture the uncertainty residing in model parameters. This entails performing Bayesian inference on the NN parameters.
The resulting probabilistic NN, a NN where each parameter has a distribution of values instead of a single value, is also known as a \emph{Bayesian neural network} (BNN)~\cite{MacKay1992, Neal1996, Graves2011, Blundell2015, Gal2016}.
The Bayesian paradigm presents a principled framework to update parameter uncertainty, from a \emph{prior} distribution to a \emph{posterior} distribution, when new data becomes available. The central task for a Bayesian problem is thus to compute the posterior parameter distribution.

A conventional approach to Bayesian inference is to draw samples from the posterior distribution using Markov chain Monte Carlo (MCMC) algorithms~\cite{Andrieu2003,Various2011}.
While MCMC methods offer desirable asymptotic convergence to the true posterior, they tend to converge slowly and do not scale well to high dimensional parameter spaces.
While advanced MCMC variants such as Hamiltonian Monte Carlo (HMC)~\cite{Duane1987,Neal1996} have been exercised for BNNs, the networks studied typically have had a few hundred parameters (e.g.,~\cite{He2012, Chen2014, Zhang2018}).

Variational inference (VI)~\cite{Blei2017,Zhang2019} is an alternative to MCMC that is more scalable.
Instead of MCMC sampling, VI looks for a best approximate posterior from a family of tractable distributions, thereby turning a sampling task into an optimization problem.
The simplest is perhaps mean-field VI~\cite{Graves2011, Blundell2015} that searches within the space of all \emph{independent} Gaussian distributions. While very scalable to high dimensions, mean-field VI is unable to capture correlation among parameters and tends to underpredict the variance~\cite{Blei2017}.
Beyond mean-field, particle-based Stein VI methods such as the Stein variational gradient descent (SVGD)~\cite{Liu2016}  algorithm has demonstrated an attractive balance between computational scalability and an ability to capture general non-Gaussian distributions.
However, SVGD has been observed to carry a tendency to collapse to the posterior mode when the particle-to-dimension ratio becomes low~\cite{Zhuo2018, Wang2018}.
More recently, the projected SVGD (pSVGD)~\cite{Chen2020} that forms a lower-dimensional posterior most informed by the data is able to increase the particle efficiency and mitigate the mode-collapse phenomenon.
Both SVGD and pSVGD have only been lightly explored for usage in constructing BNNs~\cite{Morelli2023,Hauth2024}; we will provide further study of the application of these algorithms in this paper.

We demonstrate SVGD and pSVGD on NNs of spatially complex, history dependent processes that emerge from SciML applications that are of interest in engineering and materials science.
The first is homogenization of the inelastic mechanical response of stochastic volume elements of polycrystalline metals \cite{frankel2019oligocrystals,frankel2019evolution}.
The second models the process of out-gassing of a solid-state fuel \cite{aagesen2020phase,kim2022modeling} through complex networks formed by grain boundaries.
Models such as these can be used a subgrid constitutive models in large scale simulations, efficient surrogates for  structure-property optimization and exploration, and in material variability quantification.

The key contributions of this paper are:
\begin{itemize}
\item We enable Bayesian UQ for GCNNs that emerge from two complex SciML applications of polycrystal stress response and out-gassing flux.

\item We obtain the Bayesian UQ information by implementing and deploying SVGD and pSVGD to accommodate the GCNN architecture.

\item We extract insights on the extremely narrow BNN posteriors resulting from the instability of NN model due to the underlying physics and time-integration requirements.

\end{itemize}

The paper is organized as follows.
\sref{sec:methodology} provides an overview of the HMC, SVGD, and pSVGD algorithms used for conducting Bayesian inference of the NN parameters.
\sref{sec:exemplars} provides details the two SciML applications and their GCNN architectures.
\sref{sec:results} then presents numerical results of the BNN construction from the different inference algorithms, along with the ensuing uncertainty information in the model and predictions.
The paper concludes in \sref{sec:conclusion} with a summary of the key findings and future work.

\section{Methodology} \label{sec:methodology}

NNs have seen success in representing processes with spatial-temporal complexity.
A common approach combines two components: (a) a parameterized convolution for spatial dependence, and  (b) a learnable recurrence relation for temporal evolution~\cite{shi2015convolutional,azad2019bi,lin2020self}.
Thus, an input is first reduced to a latent description through convolutions, evolved by the recurrent component, and then decoded to the output space.
For the spatial component, pixel-based convolutional neural networks (CNNs) operate on grid-structured data while graph-based CNNs (GCNNs) relax this requirement.
Graph convolutions are flexible in the sense that they work equally well on the connectivity of structured grids and unstructured meshes, or on data that has a network relationship.
For the temporal component,  recurrent neural networks (RNNs) that describe time-integrator-like recurrent relations have seen wide application.
These RNNs include specific architectures that target history-dependent processes such as the long short-term memory (LSTM) unit~\cite{hochreiter1997long} and the gated recurrent unit (GRU)~\cite{cho2014learning} that involve gating to form ``long'' and ``short''-term memory.
With similar goals, neural ordinary differential equations (NODEs)~\cite{chen2018neural,dupont2019augmented} follow the well-known discrete time integrators of scientific computing~\cite{hairer1993solvingI,wanner1996solvingII,hairer2006geometric} to learn a NN representation of the ``right-hand-side'' of ordinary differential equations (ODEs) describing the latent dynamics.

Generically, these models follow a three-part architecture:
\begin{align} \label{eq:NN}
\outputvectorP = \NN(\inputvector, \timevector; \weight) =
\begin{cases}
\hiddenvector_{0}   &= \NN_\inputvector  (\inputvector; \weight_0)  \\
\hiddenvector_{n+1} &= \NN_\hiddenvector (\hiddenvector_n,\timevector_n; \weight_n) \\
\outputvectorP_n     &= \NN_\outputvector(\hiddenvector_n; \weight_y)
\end{cases}.
\end{align}
First, the physical initial condition $\inputvector$ is encoded into an initial hidden state  $\hiddenvector_0$, via encoder $\NN_\inputvector$ that is a convolutional NN with reduction via pooling.
Then, the hidden state $\hiddenvector_n$, together with a time-like input $\timevector_n$, is evolved in a time-integrating fashion by a RNN- or NODE-based $\NN_\hiddenvector$.
Finally, each hidden states is transformed to predict the output quantity of interest (QoI) $\outputvectorP_n$ via decoder $\NN_\outputvector$, where the hat-notation indicates a \emph{prediction} from the NN model.
Collectively, we denote the entire time-trace quantities as
$\hiddenvector\coloneqq\{\hiddenvector_n\}_n$, $\timevector\coloneqq\{\timevector_n\}_n$,
and $\outputvectorP\coloneqq\{\outputvectorP_n\}_n$.
$\NN_\inputvector$, $\NN_\hiddenvector$, and $\NN_\outputvector$ are respectively parameterized by $\weight_0$, $\weight_n$, and $\weight_y$, and similarly $\weight \coloneqq \{\weight_0, \weight_y, \{\weight_n\}_n \}$ denotes the entire collection of weights for the model.

We explore $\NN_\hiddenvector$ in the forms of RNN and NODE.
An RNN architecture resembles filters from control theory. For example, the particular form of a GRU update is
\begin{eqnarray} \label{eq:rnn}
\hiddenvector_{n+1} = \NN_\hiddenvector (\hiddenvector_n,\timevector_n; \weight_n) &=&  (\Is - \zs_{n+1}) \odot \hiddenvector_{n} + \zs_{n+1} \odot \tilde{\hiddenvector}_{n+1}, \label{eq:gru}
\\
&=&   \hiddenvector_{n} + \zs_{n+1} \odot (\tilde{\hiddenvector}_{n+1} - \hiddenvector_{n} ), \nonumber
\end{eqnarray}
where $\odot$ denotes the Hadamard element-wise product, and $\tilde{\hiddenvector}_{n+1}$ is a trial state and $\zs_{n+1}$ provides a gating mechanism on the update from $\hiddenvector_n$ to $\hiddenvector_{n+1}$.
Both the gate and trial state are computed by densely connected NNs
on inputs $\timevector_n$ and $\hiddenvector_n$.
The second form in \eref{eq:rnn} illustrates a resemblance of the gated RNN update with that of a discrete time integrator.
A NODE is derived from time-integration of ODEs. For example, a forward Euler discretization of an underlying continuous-time ODE
$\dot{\hiddenvector} = \Rs (\hiddenvector,\timevector)$
leads to
\begin{align}
\hiddenvector_{n+1} &=\NN_\hiddenvector (\hiddenvector_n,\timevector_n; \weight_n) = \hiddenvector_{n} + \dt \Rs (\hiddenvector_n,\timevector_n; \weight_n),
\end{align}
where $\dt$ is the timestep.

Additional architecture details that are specific to the SciML applications will be presented in \sref{sec:exemplars}.

\subsection{Bayesian uncertainty quantification}
\label{ss:Bayesian}

A NN is trained using a training dataset of $N_{D}$ time-traces: $\data = \{ \inputvector_i, \timevector_i, \outputvector_i \}_{i=1}^{N_{D}}$.
An ordinary, non-probabilistic training involves optimizing the NN weight parameters to minimize some loss function that reflects the predictive capability of the overall NN model, for example $\weight^{\ast} = \arg\min_{\weight} \mathcal{L}(\weight; \data)$.
A common loss function is the mean-square error (MSE).
In contrast, a Bayesian (probabilistic) training of the NN, resulting in a BNN, treats the weight parameters as a random vector that has an associated probability distribution.
Given training data, Bayes' rule provides an update of the parameter distribution:
\begin{align} \label{eq:bayes}
\prob(\weight \giv \data)
= \frac{ \prob(\data \giv \weight) \, \prior(\weight)}{\prob(\data)} = \frac{ \prob(\{\outputvector_i\}_i \giv \{\inputvector_i,\timevector_i\}_i, \weight) \, \prior(\weight)}{\prob(\{\outputvector_i\}_i)},
\end{align}
where $\prior(\weight)$ is the \emph{prior} probability density that represents the uncertainty of NN parameters prior to considering the training data,
$\prob(\weight \giv \data)$ is the \emph{posterior} probability density that represents the updated uncertainty of NN parameters after conditioning on data,
$\prob(\data \giv \weight)$
is the \emph{likelihood} that describes the probability of observing the data under a given parameter setting $\weight$, and
$\prob(\data)$ is the marginal likelihood (also called the model evidence) that does not depend on the parameters $\weight$.
In the context of NNs, an uninformative independent normal distribution $\prior(\weight)\sim \Nc(\mathbf{0},\sigma_0^2 \Ib)$  is often adopted for the prior since many of the weights are fungible and have complex relationships with the physical outputs.

More precisely, the likelihood that is actually being computed is  $\prob(\{\outputvector_i\}_i \giv \{\inputvector_i,\timevector_i\}_i, \weight)$.
It is usually established following an observation model (for all $\outputvector_n$ in each time-trace) in the form $\outputvector_{i,n} = \outputvectorP_{i,n}(\inputvector_i,\timevector_i,\weight) + \epsilon_{i,n}$ where $\epsilon_{i,n} \sim \mathcal{N}(0,\sigma_{\epsilon}^2)$ is an independent Gaussian discrepancy with  $\sigma_{\epsilon}$ reflecting the level of data noise.
The independence of $\epsilon_{i,n}$ allows the joint likelihood to be decomposable into product of marginals: $\prob(\{\outputvector_i\}_i \giv \{\inputvector_i,\timevector_i\}_i, \weight)= \prod_{i,n} \prob(\outputvector_{i,n} \giv \inputvector_i,\timevector_i, \weight)$.

We will introduce the specific methods for computing the posterior $\prob(\weight \giv \data)$ in subsections below. Once the posterior is obtained, one can then use the resulting BNN to obtain a corresponding distribution of predictions under given new input $\inputvector^{\text{new}}$ and $\timevector^{\text{new}}$, through the \emph{posterior push-forward} distribution
\begin{align}
\prob(\hat{\outputvector}^{\text{new}} \giv \inputvector^{\text{new}},\timevector^{\text{new}},\data)=\prob(\NN(\inputvector^{\text{new}},\timevector^{\text{new}}; \weight)  \giv \inputvector^{\text{new}},\timevector^{\text{new}},\data)
\end{align}
and the \emph{posterior-predictive} distribution
\begin{align}
\prob(\outputvector^{\text{new}} \giv \inputvector^{\text{new}},\timevector^{\text{new}},\data) = \int \prob(\outputvector^{\text{new}} \giv \inputvector^{\text{new}},\timevector^{\text{new}},\weight) \,\prob(\weight \giv \data) \, d\weight.
\end{align}
The key difference is that the push-forward distribution simply propagates the parameter uncertainty through the NN and provides a distribution of NN predictions $\hat{\outputvector}$, while the predictive distribution also incorporates data noise in its overall uncertainty (e.g., the posterior predictive for the additive Gaussian observation model case is $\prob({\outputvector}^{\text{new}} \giv \inputvector^{\text{new}},\timevector^{\text{new}},\data)=\prob(\NN(\inputvector^{\text{new}},\timevector^{\text{new}}; \weight)+\epsilon  \giv \inputvector^{\text{new}},\timevector^{\text{new}},\data)
$).
Both distributions can be sampled by, for example, first generating samples $\weight_{(i)}$ from the posterior $\prob(\weight \giv \data)$, evaluating the NN model for each sample to obtain a corresponding sample of $\hat{\outputvector}^{\text{new}}_{(i)}$, and then adding a noise sample $\epsilon_{(i)}$ to obtain a sample of ${\outputvector}^{\text{new}}_{(i)}$. In this work, we will mostly study posterior push-forward behavior.

\subsubsection{Hamiltonian Monte Carlo}

Hamiltonian Monte Carlo (HMC)~\cite{Duane1987,Neal1996} is an effective sampling algorithm in the Markov Chain Monte Carlo (MCMC) family.
HMC frames the posterior sampling in terms of Hamiltonian dynamics (HD) evolution of the weights coupled with periodic Metropolis-Hastings (MH) steps~\cite{betancourt2017conceptual}.
HD allows the weights to transition away from the previous accepted state while staying in regions of high posterior probability,
as a result of phase volume-preserving properties of HD.
The usual MH acceptance criterion can then be applied to these decorrelated samples.
The overall algorithm has some similarity to the thermostatted dynamics of single particle in a high dimensional space.

The Hamiltonian is comprised of a potential and a kinetic energy term:
\begin{align}
\hamiltonian = \potential + \frac{1}{2} \momentum \cdot \massmatrix \momentum.
\end{align}
The posterior $\prob(\weight \giv \data)$ is associated with the potential $\potential$ via
\begin{equation} \label{eq:HMC_potential}
\potential = -\log \prob(\weight \giv \data)
\end{equation}
and the (fictitious) momentum $\momentum$ is combined with a selected mass matrix $\massmatrix$ to form a quadratic kinetic energy.
The choice of the mass matrix effectively transforms the target parameter space by changing the constant $\hamiltonian$ level sets and the correlation structure of the target distribution in terms of $(\weight,\momentum)$.
The choice of the form of kinetic energy is relatively free since the main requirement is that the probability distribution on the phase space marginalizes to the target distribution.

Computationally, HD is implemented with a sympletic (reversible, phase volume-preserving) integrator such as leapfrog Verlet:
\begin{align}
\momentum_k ((\tilde{n}+1/2)\dt)   &= \momentum_k(\tilde{n} \dt) - \frac{1}{2} \dt \grad \potential(\weight_k) \nonumber \\
\weight_k ((\tilde{n}+1) \dt)   &= \weight_k (\tilde{n} \dt) +  \dt \massmatrix^{-1} \momentum_k((\tilde{n}+1/2) \dt) \label{eq:HD} \\
\momentum_k((\tilde{n}+1)\dt)    &= \momentum_k((\tilde{n}+1/2) \dt) - \frac{1}{2} \dt \grad \potential(\weight_k ((\tilde{n}+1) \dt), \nonumber
\end{align}
where $k$ indexes the HD stage and $\tilde{n}$ indexes pseudo-time of the HD trajectory.
The time-integration generally requires the gradient of the log posterior with respect to the weights, which can make HMC computationally expensive relative to simpler MC methods.
The effect of the mass matrix $\massmatrix$ is evident from how it enters \eref{eq:HD}, by transforming $\momentum_k$ and the effect of $\grad \Phi$ on $\weight_k$.
The pseudo-time step $\dt$ is chosen to maintain dynamic stability while promoting efficient decorrelation of states.
The momentum provides a stochastic contribution to the overall dynamics: the initial momentum at stage $k$ is a blend of the momentum from the previous HD stage plus a random component,
\begin{align}
\momentum_k(0) = \beta \momentum_{k-1} + (1-\beta) \varepsilonb \qquad \text{with} \qquad \varepsilonb \sim \Nc(\mathbf{0},\massmatrix),
\end{align}
where $\beta$ is a hyperparameter and $\momentum_0 = \varepsilonb$ for the first stage.

Lastly, the MH step for accepting the sample at each HMC stage $k$ is carried out after integrating $N_{\text{MH}}$ pseudo-time steps:
\begin{align}
\weight_{k+1} \giv \weight_k =
\begin{cases}
\weight_k(N_{\text{MH}} \dt) & \text{with probability} \ \alpha \\
\weight_k(0)     & \text{else}
\end{cases},
\end{align}
where the acceptance probability of the time-integrated value $\weight_k(N_{\text{MH}} \dt)$  is
\begin{align}
\alpha = \min \left( 1, \frac{\exp(- \hamiltonian(N_{\text{MH}} \dt) }{ \exp(- \hamiltonian(0) )} \right).
\end{align}
The MH and the momentum initialization allow the overall chain to jump to new level sets of the Hamiltonian.

\subsubsection{Stein variational gradient descent}

Here we briefly review SVGD~\cite{Liu2016} and pSVGD~\cite{Chen2020} methods that provide a coordinated ensemble of parameter samples through a type of functional gradient descent.

The methods are based on the Stein identity
\begin{align}
\EX_{\xb\sim\pi}[ \Ac_\prob [ \phib(\xb) ] ] = \mathbf{0}
\end{align}
for a smooth function $\phib$, where the operator $\Ac_\prob$ is
\begin{align}
\Ac_\prob [\phib(\xb)] = \grad_\xb [ \log \prob(\xb)] \otimes \phib(\xb) + \grad_\xb \phib(\xb)
\end{align}
for a generic probability density $\pdf(\xb)$.
Notably, the Stein operator $\Ac_{\prob}$ depends on the density $\prob$ only through the \emph{score} function $\grad_\xb [\log \prob(\xb)]$ which is independent of the normalization of $\pdf$ and thus is suitable for Bayesian inference where only the unnormalized posterior is available (i.e., having only access to prior times likelihood).
If we let $\tilde{\prob}$ be an approximate density under a perturbation mapping of the random variable $\xb \mapsto \xb+\epsilon \phib$, then the expectation of $\tr (\Ac_\prob [ \phib (\xb)])$ can be shown to relate to the gradient of the Kullback--Leibler (KL) divergence between $\tilde{\prob}$ and $\prob$~\cite[Thm 3.1]{Liu2016}:
\begin{align}
\left.
\grad_\epsilon D_{\text{KL}}( \tilde{\prob}_{\xb+\epsilon \phib} ||  \prob)
\right|_{\epsilon=0}
= - \EX_{\tilde{\prob}} [ \tr(\Ac_\prob [ \phib(\xb) ] ].
\end{align}
This connection provides the basis for a gradient-based method to update the approximate posterior density to the true posterior density.

Leveraging these relationships and with some additional derivations (see~\cref{Liu2016} for details), the SVGD algorithm initializes an ensemble of $N_p$ particles $\{\weight_{(i)}\}_{i=1}^{N_p}$ in the parameter space and updates them via gradient-descent steps
\begin{align}
\weight^{k+1}_{(i)} = \weight^k_{(i)} + \gamma \gs({\weight^k_{(i)}}) \ ,
\end{align}
where $\gamma$ is a learning rate parameter and
\begin{align}    \label{eq:stein_grad}
\gs(\weight) = \frac{1}{N_p}\sum_{j=1}^{N_p} \kernel(\weight^k_{(j)}, \weight) \grad_{\weight^k_{(j)}} [\log \posterior(\weight^k_{(j)} \giv \data) ]+  \grad_{\weight^k_{(j)}} \kernel(\weight^k_{(j)}, \weight)
\end{align}
is a gradient estimate.
The kernel $\kernel(\cdot,\cdot)$ is required to have certain regularity conditions, which can be satisfied by the radial basis function (RBF) kernel $\kernel(\weight,\weight')=\exp(-\frac{1}{h}||\weight-\weight'||^2)$ for example~\cite{Liu2016}.
Discussions on additional choices of the kernel, as well as incorporating gradient and approximate Hessian information into the kernel construction, can be found  in~\cref{wang2019stein}.
\eref{eq:stein_grad} can be interpreted as a kernel-smoothed gradient that drives the particles toward the high-probability region of the posterior, and a repulsion term that keeps the particles separate.
At convergence, the particles collectively approximate the posterior distribution.

\subsubsection{Projected Stein variational gradient descent}

SVGD has been observed to have a tendency for the particles to collapsing toward the maximum a posteriori (MAP) point (i.e., posterior mode) when the dimension of the parameter space, $N_w$, is much greater than the number of particles, $N_p$~\cite{Zhuo2018}, which in turn underestimates the uncertainty in these situations.
pSVGD~\cite{Chen2020} combats this phenomenon by seeking to perform Bayesian inference only a lower-dimensional active subspace of the full parameter space.
This subspace is based on the likelihood Hessian
\begin{align}
\Hs = \grad_\weight \left[ \grad_\weight \pdf(\outputvector \giv \inputvector,\timevector,\weight)  \right]
\end{align}
at the MAP,  either in its full form or its Gauss-Newton first derivative approximation.
The subspace is constructed from the $r$ most significant generalized eigenvectors obtained from
\begin{align}
\Hs \psib_{l} = \lambda_{l} \Sigmab_0^{-1}\psib_{l},
\end{align}
where $\Sigmab_0$ is the prior covariance.
The potentially large dimenson eigenvalue problem can be efficiently solved by randomized numerical linear algebra \cite{halko2011finding}.
The column matrix of the leading $r$ eigenvectors, $\Psib = [\psib_0, \ldots, \psib_r ]$, allows the projection of parameters into the active low-dimensional subspace where the likelihood (data) highly informs the posterior:
\begin{align}
\weight
&= \weight - \bar{\weight} + \bar{\weight} \\
&= \underbrace{\Ps [ \weight - \bar{\weight} ]}_{\weight^r}
+ \underbrace{\Qs [ \weight - \bar{\weight} ] + \bar{\weight}}_{\weight^\perp}, \nonumber
\end{align}
where $\weight^r$ is in the active subspace component, $\weight^\perp$ is the inactive compliment where the prior is dominant, and $\bar \weight$ is the particle mean  (or, alternatively, the MAP).
Hence, $\Ps = \Psib \Psib^{\top}$ is the projector to the active space and $\Qs=\Is-\Ps$ is its complement..

The weights in the inactive subspace, which are considered as the unmodified prior's projections, can be frozen at the samples of the prior.
Then, the pSVGD algorithm needs only to determine the posterior distribution in the lower-dimensional active subspace.
Bayes' theorem in the split parameter space becomes
\begin{align} \label{eq:split_bayes}
\prob(\weight \giv \data) & \propto \prob(\{\outputvector_i\}  \giv \{\inputvector_i, \timevector_i\}_i, \weight) \, \prob_0(\weight) \\
& \approx \prob(\{\outputvector_i\}  \giv \{\inputvector_i, \timevector_i\}_i, \weight^r) \, \prob_0^r(\weight^r)\prob_0^{\perp}(\weight^{\perp} )
\nonumber
\end{align}
where we assume an independent prior for the parameter subspaces, i.e., $\prob_0=\prob_0^r\,\prob_0^{\perp}$.
The log-posterior gradient in \eref{eq:stein_grad} is now replaced by the projected gradient
\begin{align}\label{eq:projected_update}
\grad_{\weight^r_{(j)}}\log \prob(\weight^r_{(j)} \giv \data) = \Psib^{\top} \grad_{\weight_{(j)}}\log \prob(\weight_{(j)} \giv \data).
\end{align}
The full-dimensional posterior is then reconstructed by mapping the low-dimensional samples $\weight^r_{(i)}$  back to its low-rank form and recombining it with the inert parts of the prior to yield a full rank approximation:
\begin{align}\label{eq:projected_reconstruction}
\weight_{(i)} = \Psib \weight^r_{(i)} + \bar \weight + \weight_{(i)}^{\perp} \ .
\end{align}

\subsection{Illustrative comparisons of HMC, SVGD, and pSVGD} \label{sec:illustration}

We conduct two simple benchmark tests to compare HMC, SVGD, and pSVGD.
In both cases the likelihood associated with the data informing the third parameter component $\weight_3$ has  high noise/low precision, and hence does not nform $\weight_3$.
Thus the posterior in that dimension remains close to the prior.

\begin{figure}[htb!]
\centering
\includegraphics[width=0.24\textwidth,trim={7cm 3cm 7cm 4.5cm},clip]{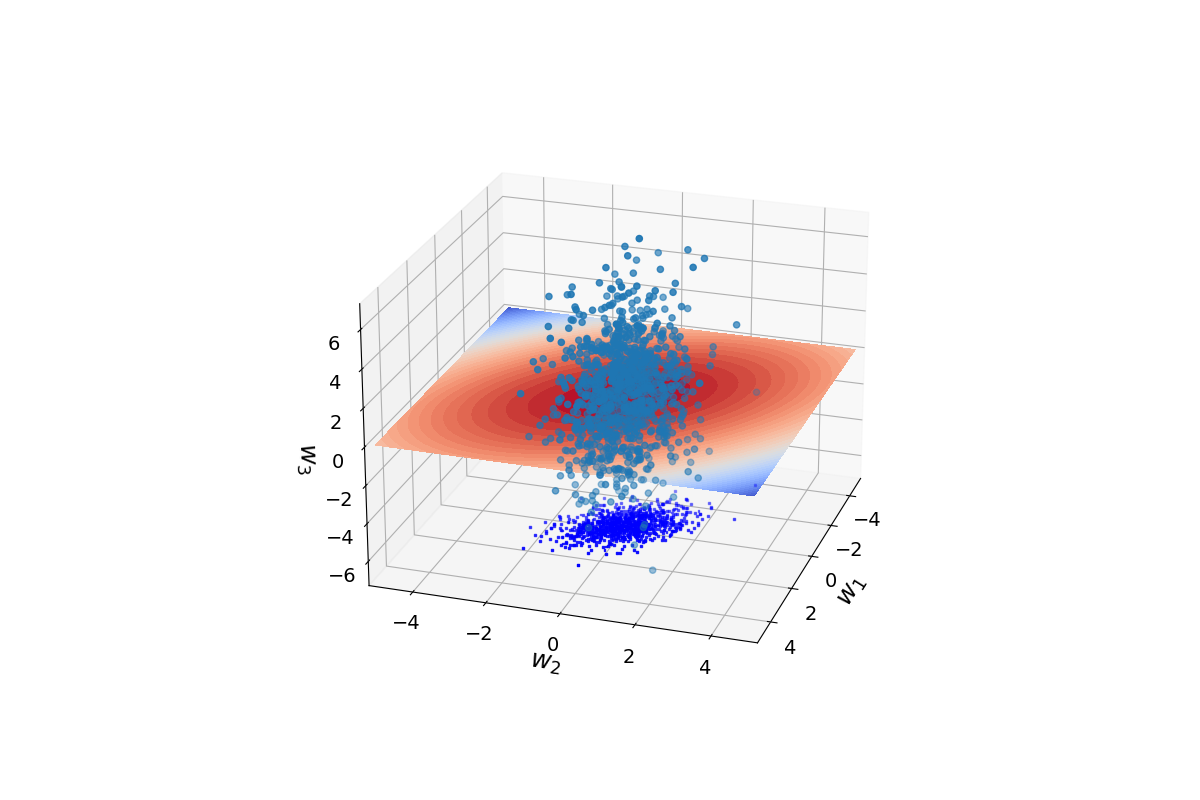}
\includegraphics[width=0.24\textwidth,trim={7cm 3cm 7cm 4.5cm},clip]{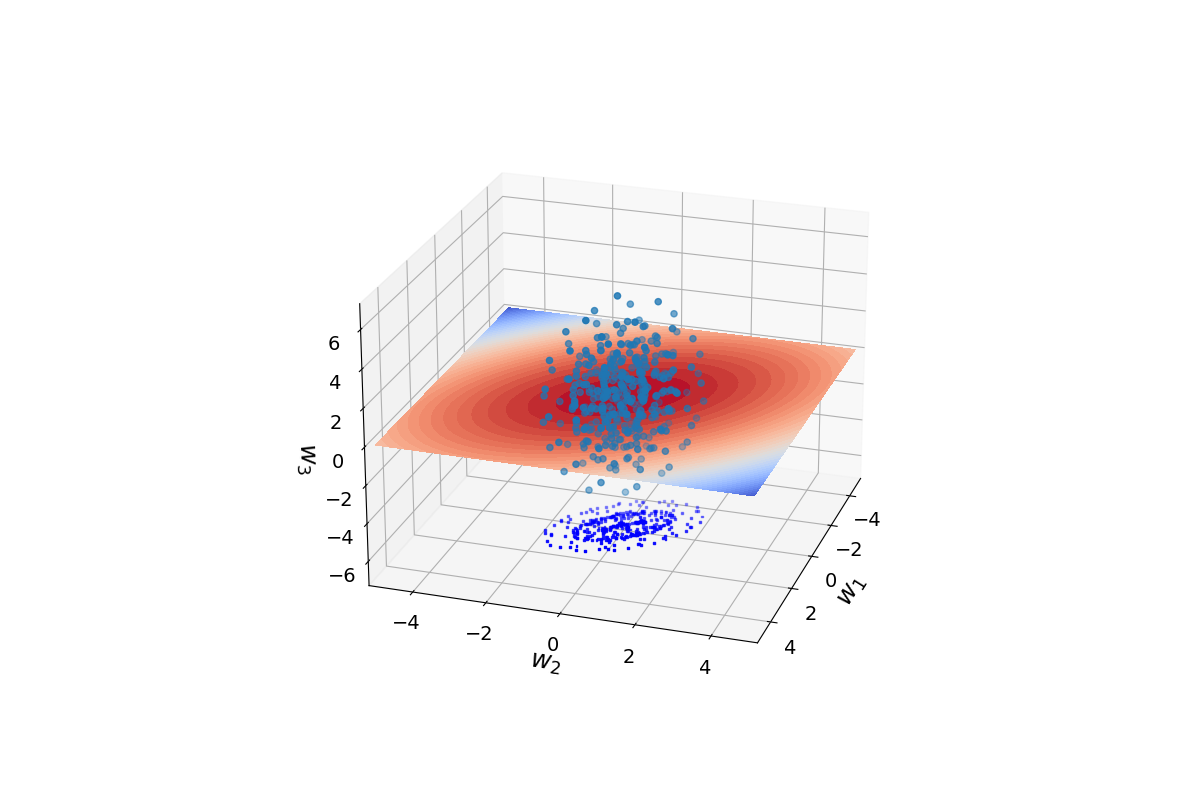}
\includegraphics[width=0.24\textwidth,trim={7cm 3cm 7cm 4.5cm},clip]{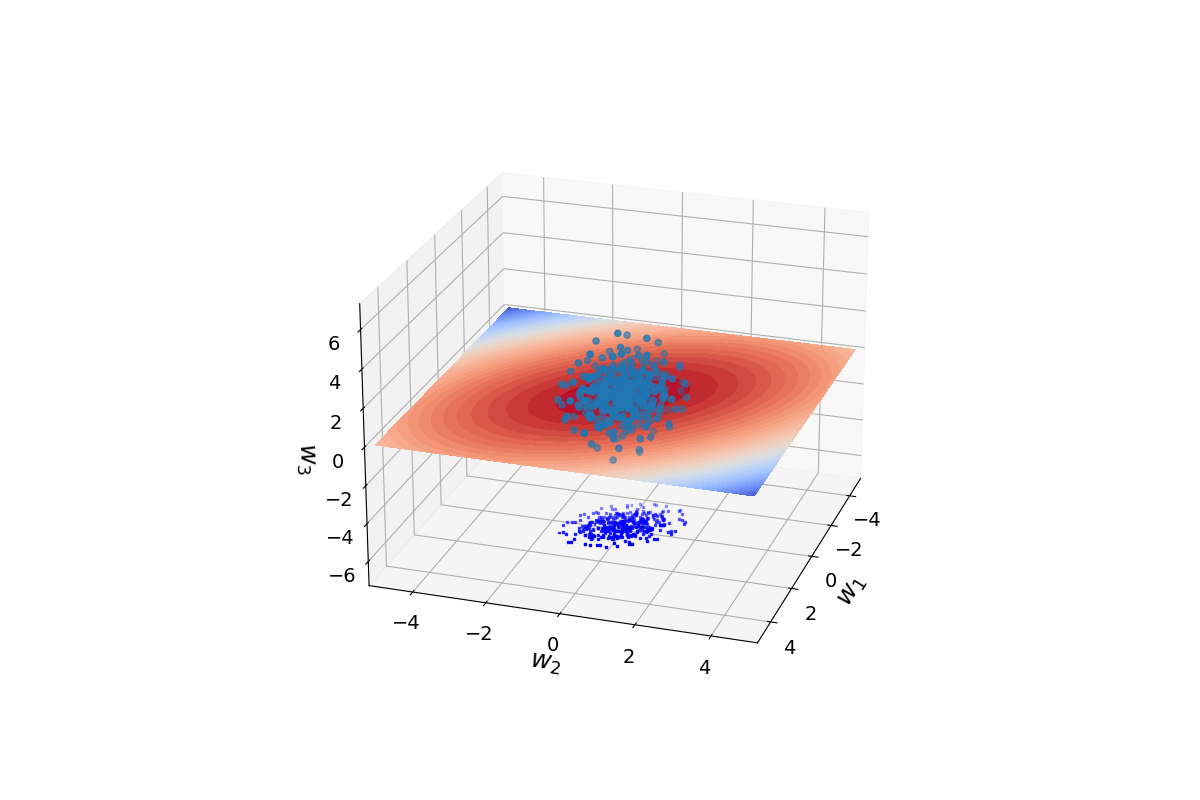}
\includegraphics[width=0.2\textwidth]{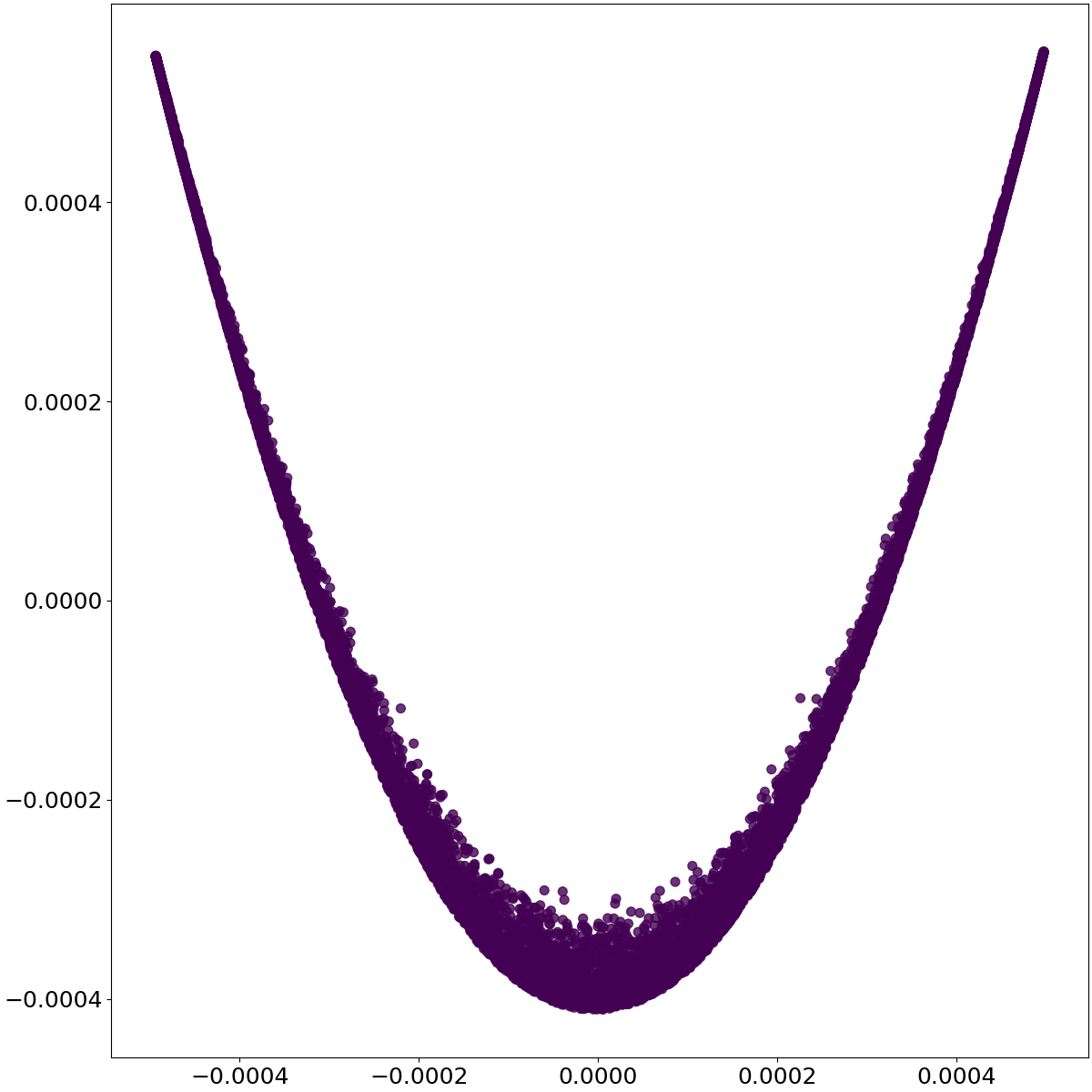}

\includegraphics[width=0.24\textwidth,trim={7cm 3cm 7cm 4.5cm},clip]{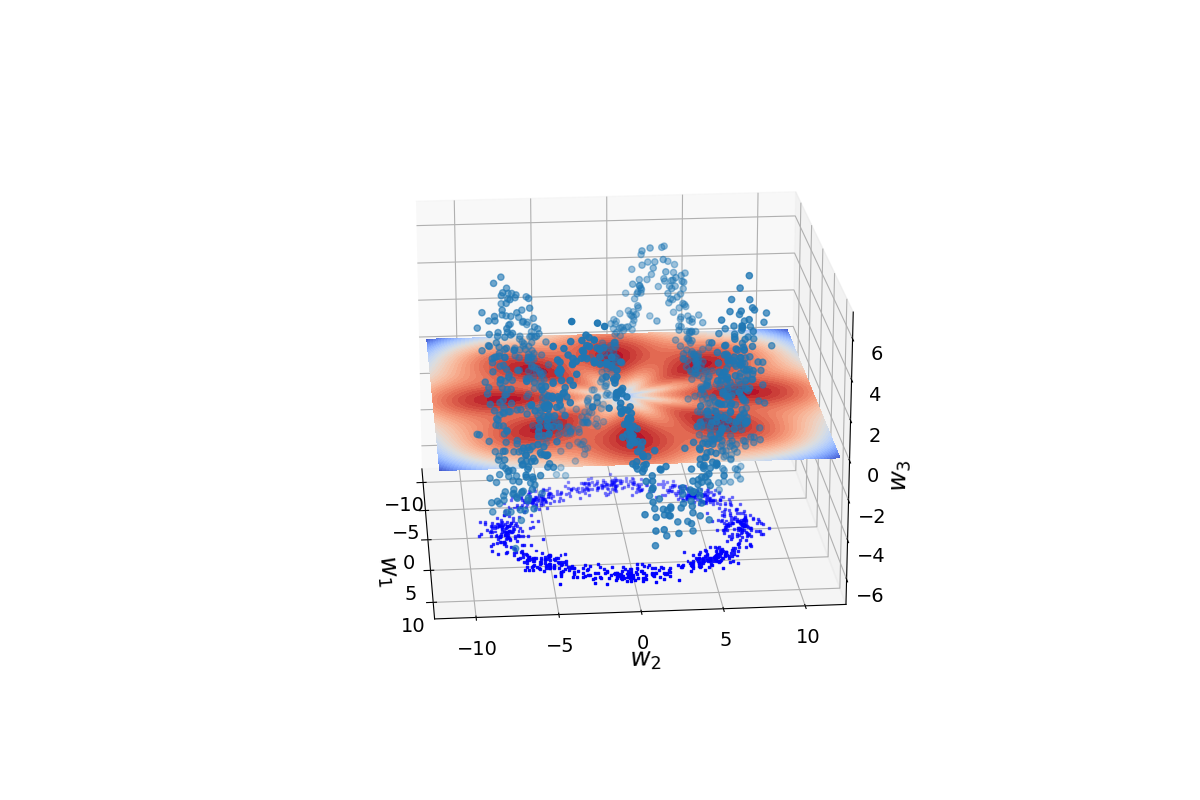}
\includegraphics[width=0.24\textwidth,trim={7cm 3cm 7cm 4.5cm},clip]{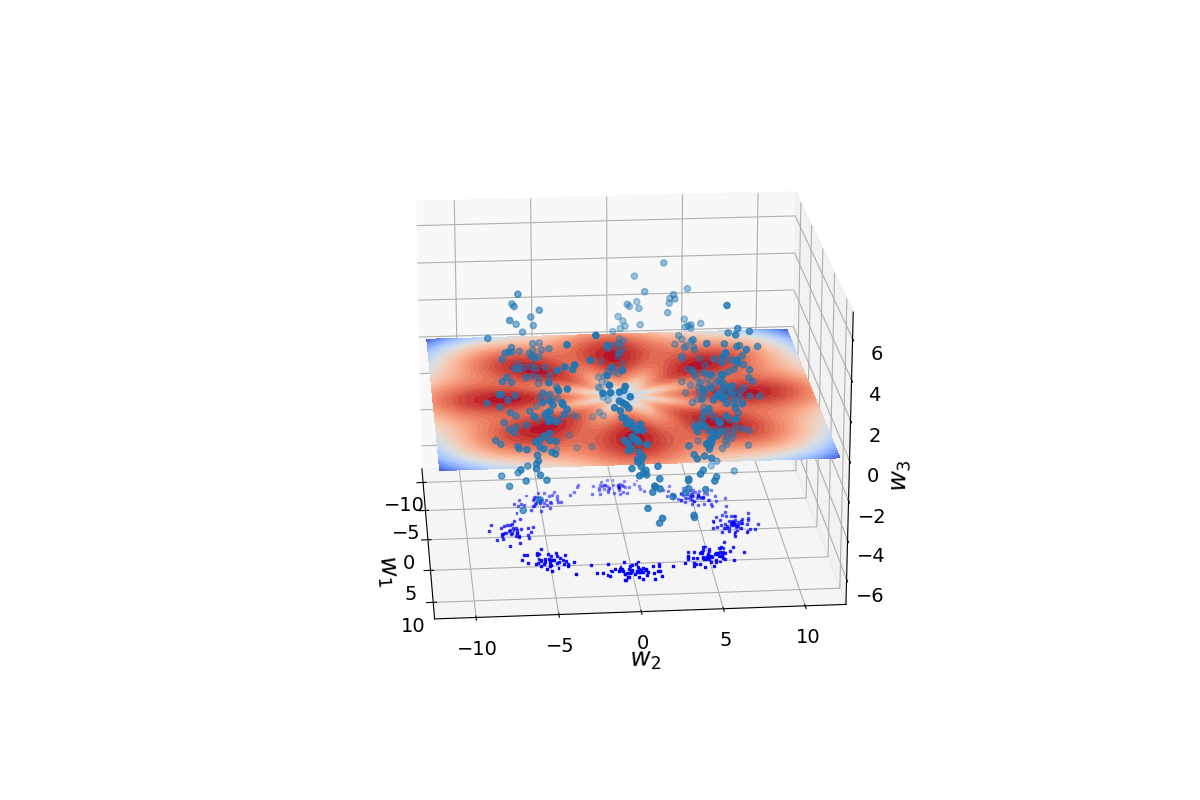}
\includegraphics[width=0.24\textwidth,trim={7cm 3cm 7cm 4.5cm},clip]{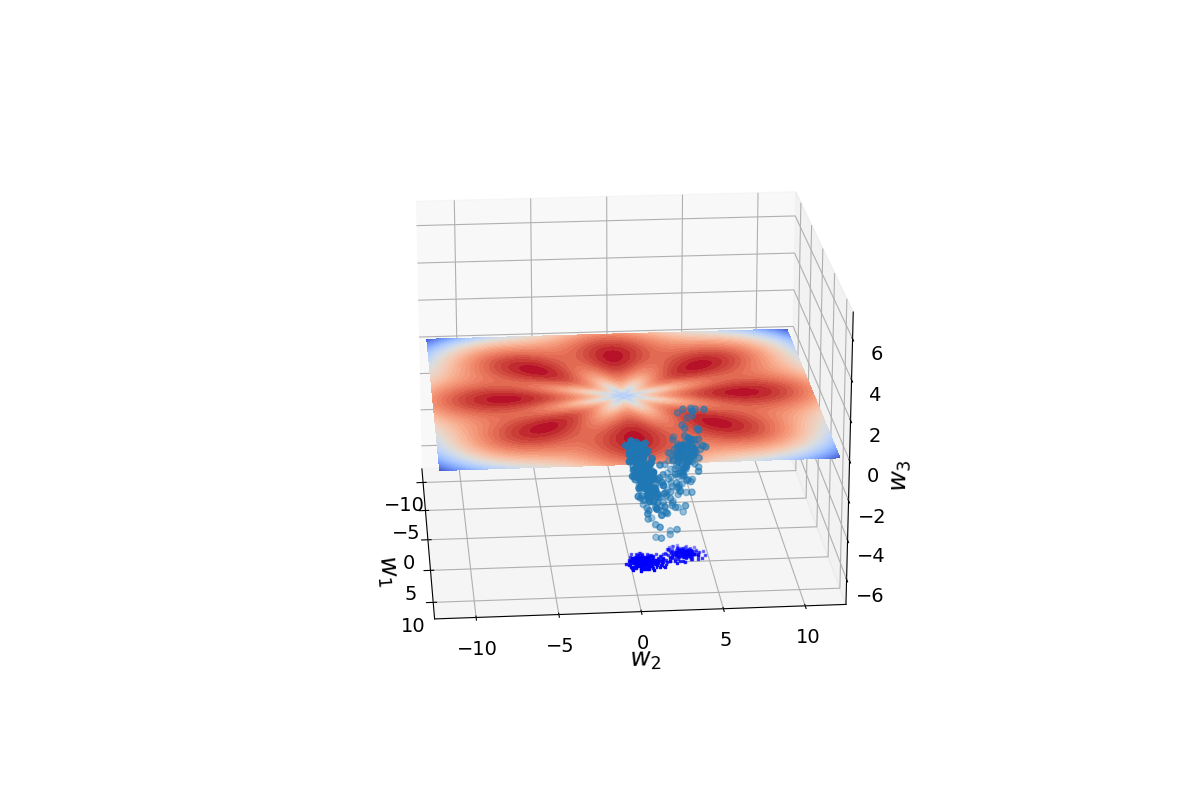}
\includegraphics[width=0.2\textwidth]{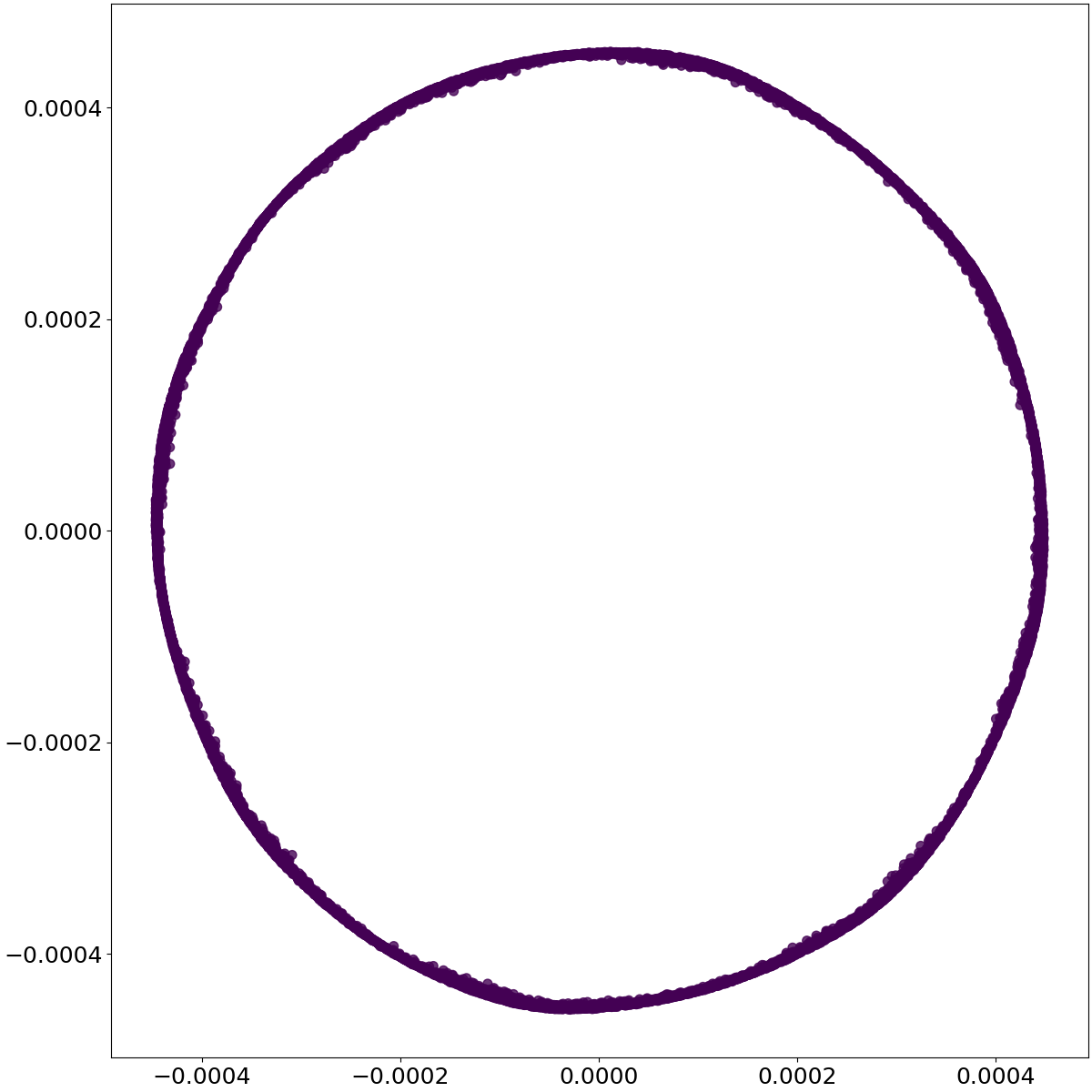}

\caption{Illustration of likelihood and samples from  HMC, SVGD, and pSVGD for MVN (upper row) and Tor (lower row).
The base shows samples projected to the $w_1$-$w_2$ plane.
Left to right: HMC, SVGD, pSVGD, diffusion map eigenvectors.
}
\label{fig:likelihood_illustration}
\end{figure}

The first example involves a 3-dimensional multivariate Gaussian posterior illustrated in \fref{fig:likelihood_illustration}.  .
pSVGD thus truncates the last parameter dimension $w_3$ and only performs inference on the two informed active directions $w_1$-$w_2$.
This example illustrates that for a well-behaved/convex posterior the SVGD samples are more structured than HMC having outer shells influenced by the repulsive part of the Stein potential and an inner density region dominated by the posterior density.
The projected SVGD largely resemble those of SVGD, albeit with less structure/variance in the less precise $w_3$ dimension.
All methods show convergence of their sample statistics to the those of the true posterior.
The results in the right frame represent the first two diffusion map~\cite{Coifman:2006,Soize:2016} eigenvectors constructed with the HMC samples.
The wider region in the center correspond the bulk of the samples near the center of the distribution while the side tails correspond to the narrowing of the distribution towards the positive/negative $w_3$ values where the spread of $(w_1,w_2)$ pairs shrinks in absolute sense.

The second example entails a highly non-Gaussian 3-dimensional posterior, as visualized in \fref{fig:likelihood_illustration} using samples/particles from the three algorithms.
This posterior shows a more complex connected set of high probability samples with adjacent MAPs connected by high density bridges, which we believe is qualitatively similar to that generated by nonlinear NNs with fungible weights.
Clearly this is illustrative of strong non-linear correlation of the parameters.
HMC and full SVGD represent the posterior well, with SVGD's fewer samples spanning the likely parameters efficiently.
Projected SVGD appears to be hampered by decoupling the dominant modes from the less dominant mode and is stuck sampling one of the many high probability posterior regions.
Similarly, the right frame displays the diffusion map eigenvectors corresponding to the HMC results for this test case.
The strong dependencies between all model parameters result in samples embedded in a circular manifold.

In this paper we will refer to possible multiple local maxima posterior distribution values as \emph{local MAPs}, to keep the terminology concise.
While this nomenclature can, in general, be confusing, in the context of this paper it is clear that we refer to local, possibly degenerate maxima of the posterior.

\section{Scientific machine learning applications} \label{sec:exemplars}

We will demonstrate building BNNs using the methods in \sref{sec:methodology} on two SciML applications featuring spatial-temporal complexity: (a) prediction of the stress response of polycrystals undergoing plastic deformation \cite{frankel2022mesh} modeled with a GCNN-RNN, and (b) prediction of the out-gassing flux of polycrystals with gas generation modeled with a GCNN-NODE.
The underlying (data-generating) models are solutions to conservation equations on a regular domain and have a single QoI that evolves in time.
We first describe the application background, data generation, and NN architecture in this section, and then present the BNN results in \sref{sec:results}.

\subsection{Mechanical response of polycrystals} \label{sec:plasticity}
The deformation response of metal polycrystals underlies the phenomenological plasticity models that enable metal forming and crash analysis.
Polycrystalline aggregates, where crystal plasticity governs the individual grain response, are a common target for homogenization in pursuit of sub-grid and structure-property surrogate models~\cite{de2019localization,frankel2020prediction,vlassis2021sobolev,de2022predicting,frankel2022mesh,vlassis2023geometric}.
These models help predict plastic deformation that can lead to structural failure.

\subsubsection{Physical system} \label{sec:cp}
Each polycrystalline sample is a collection of convex sub-regions  $\Omega_K$ of a square $\Omega_\square$, and the number of $\Omega_K$ comprising the domain $\Omega_\square$ vary sample to sample.
Each sub-region represents an individual crystal in the sample aggregate.
The crystal lattice orientation distinguishes one crystal from another, which is characterized by the angle field $\phi(\Xb)$, where $\Xb$ is the position vector in the initial configuration of the polycrystal.
For this exemplar, the angles are drawn from a uniform distribution, $\Uc[0,2\pi]$.

In this application, the QoI is the time-varying average stress $\bar{\stress}(t)$, which depends on deformation history.
Crystal plasticity models \cite{dawson2000computational,roters2010overview} are comprised of an algebraic relation mapping recoverable elastic strain to stress and a set of ODEs governing the evolution of irrecoverable plastic strain.
The stress field $\stress = \stress(\Xb,t)$ is given by
\begin{equation}
\stress = \stress(\strain_e) =  \Cbb : \strain_e,
\end{equation}
where $\Cbb$ is the 4th order elastic modulus tensor and  $\strain_e$ is elastic strain.
In each sub-region $\Omega_K$, the orientation vector $\phi$ rotates $\Cbb = \Rb(\phi) \boxtimes \underline{\Cbb}$ from the canonical $\underline{\Cbb}$.
Here $\boxtimes$ is the Kronecker product:
\begin{equation}
\Rb \boxtimes [\Ab]_{ij\ldots} \eb_i \otimes \eb_j \otimes \ldots = [\Ab]_{ij\ldots} \Rb \eb_i \otimes \Rb \eb_j \otimes \ldots.
\end{equation}
We chose the common face centered cubic (FCC) symmetry for each crystal, which has three independent components of the elastic modulus tensor $\Cbb$.
In each crystal, plastic flow can occur on any of the 12 FCC slip planes defined by slip plane normals $\nb_\alpha$ and slip directions $\sb_\alpha$, where $\sb_\alpha \perp \nb_\alpha$.
The plastic velocity gradient, which determines the plastic strain
\begin{equation}
\Lb_p = \Lb_p(\Sb) = \sum_\alpha \dot{\gamma}_{\alpha} (\Sb) \, \sb_\alpha \otimes \nb_\alpha
\end{equation}
is governed by slip rates  $\dot{\gamma}_{\alpha}$, the rotated dyads $\sb_\alpha \otimes \nb_\alpha  = \Rb(\phi) \boxtimes \left( \underline{\sb}_\alpha \otimes \underline{\nb}_\alpha \right)$, and the stress $\Sb$ at the particular location.
The slip rate $\dot{\gamma}_{\alpha}$ follows a power law
\begin{equation}
\dot{\gamma}_{\alpha}=\dot{\gamma}_0\left|\frac{\tau_{\alpha}}{g_{\alpha}}\right|^{m-1}\tau_{\alpha}
\end{equation}
driven by the shear stress $\tau_\alpha = \sb_\alpha \cdot \Sb \nb_\alpha$ resolved on slip system $\alpha$,
while the slip resistance $\dot{g}_\alpha$ evolves according to~\cite{Kocks1976, mecking1976hardening}:
\begin{equation}
\dot{g}_\alpha = (H-R_d g_\alpha) \sum_\alpha |\dot{\gamma}_\alpha| .
\end{equation}
\tref{tab:cp_parameters} summarizes parameter descriptions and values.

\begin{table}[htb!]
\centering
\begin{tabular}{|l|cl|}
\hline
Elastic moduli  & $C_{11}$ &  1.000 \\
& $C_{12}$ &  0.673 \\
& $C_{12}$ &  0.617 \\
\hline
Hardening modulus & $H$ &     0.00174\\
Initial slip resistance & $g_{\alpha}$ & 0.000597  \\
\hline
Applied strain rate & $\dot{\bar{\epsilon}}$ & 1.000 \\
Reference slip rate & $\dot{\gamma}_0$ & 1.000 \\
\hline
Rate sensitivity exponent & $m$ & 20 \\
Recovery constant & $R_d$       & 2.9  \\
\hline
\end{tabular}
\caption{Crystal plasticity parameters representative of steel.
All moduli are made dimensionless by $C_{11}=$ 205~GPa and all rates were non-dimensionalized by the the applied strain rate $1 s^{-1}$.
The maximum strain attained by the simulations was $0.003$.
Note the moduli $[\Cbb]_{iiii} = C_{11}$, $[\Cbb]_{iijj} = C_{12}$, and $[\Cbb]_{ijij} = C_{44}$, where $i\neq j$. }
\label{tab:cp_parameters}
\end{table}

\begin{figure}[htb!]
\centering
\begin{subfigure}[b]{0.30\textwidth}
\centering
\includegraphics[width=1.00\textwidth]{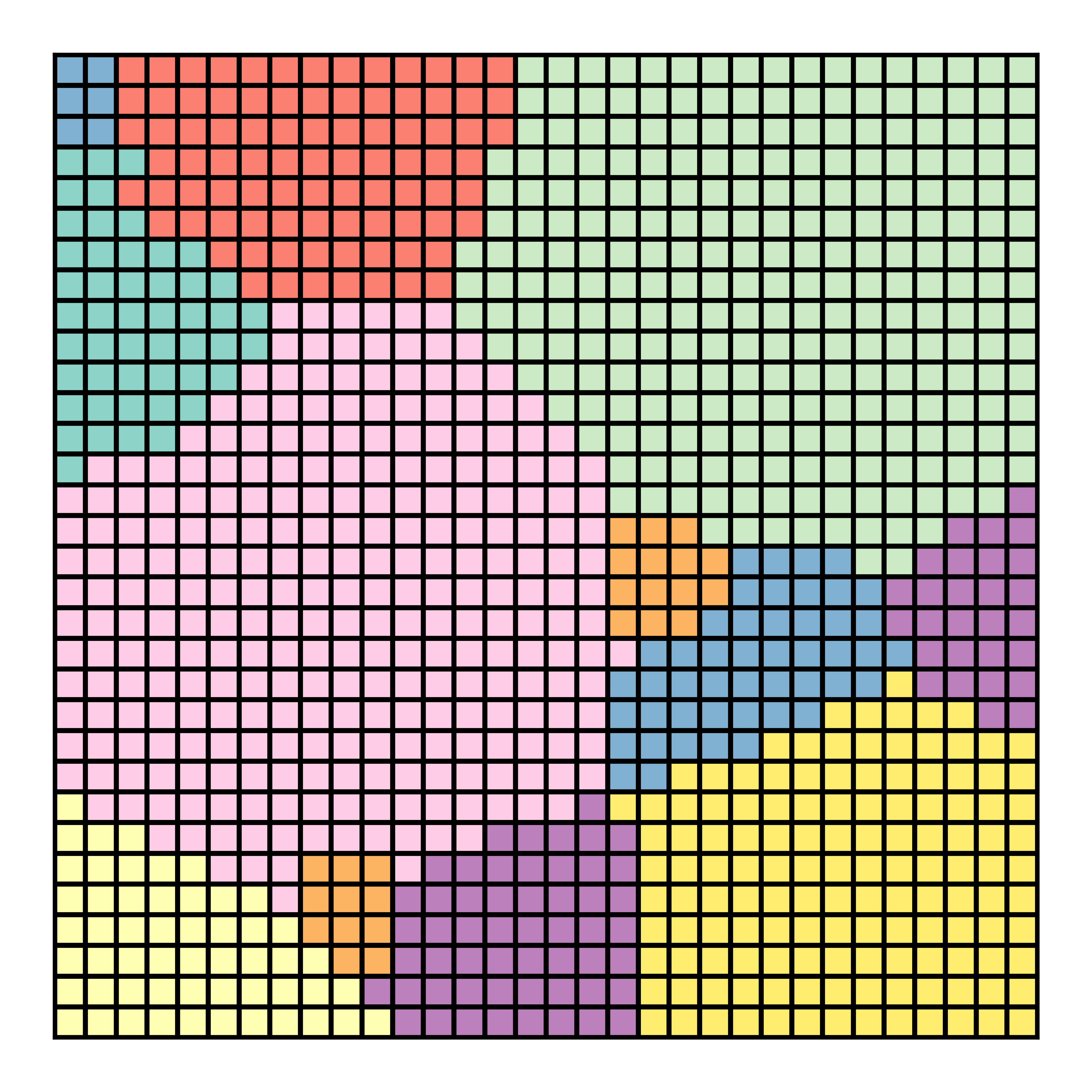}
\end{subfigure}
\begin{subfigure}[b]{0.30\textwidth}
\centering
\includegraphics[width=1.00\textwidth]{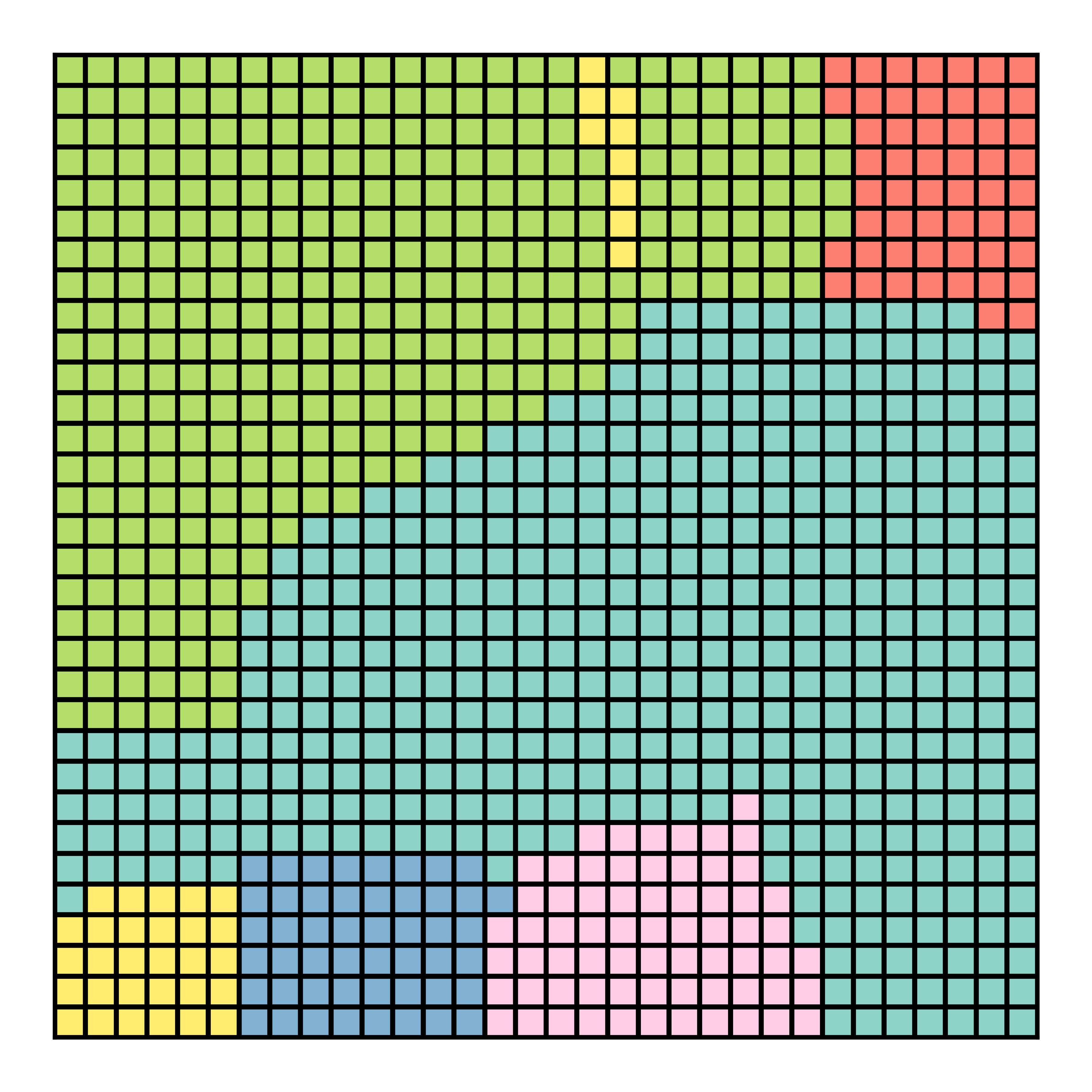}
\end{subfigure}
\begin{subfigure}[b]{0.30\textwidth}
\centering
\includegraphics[width=1.00\textwidth]{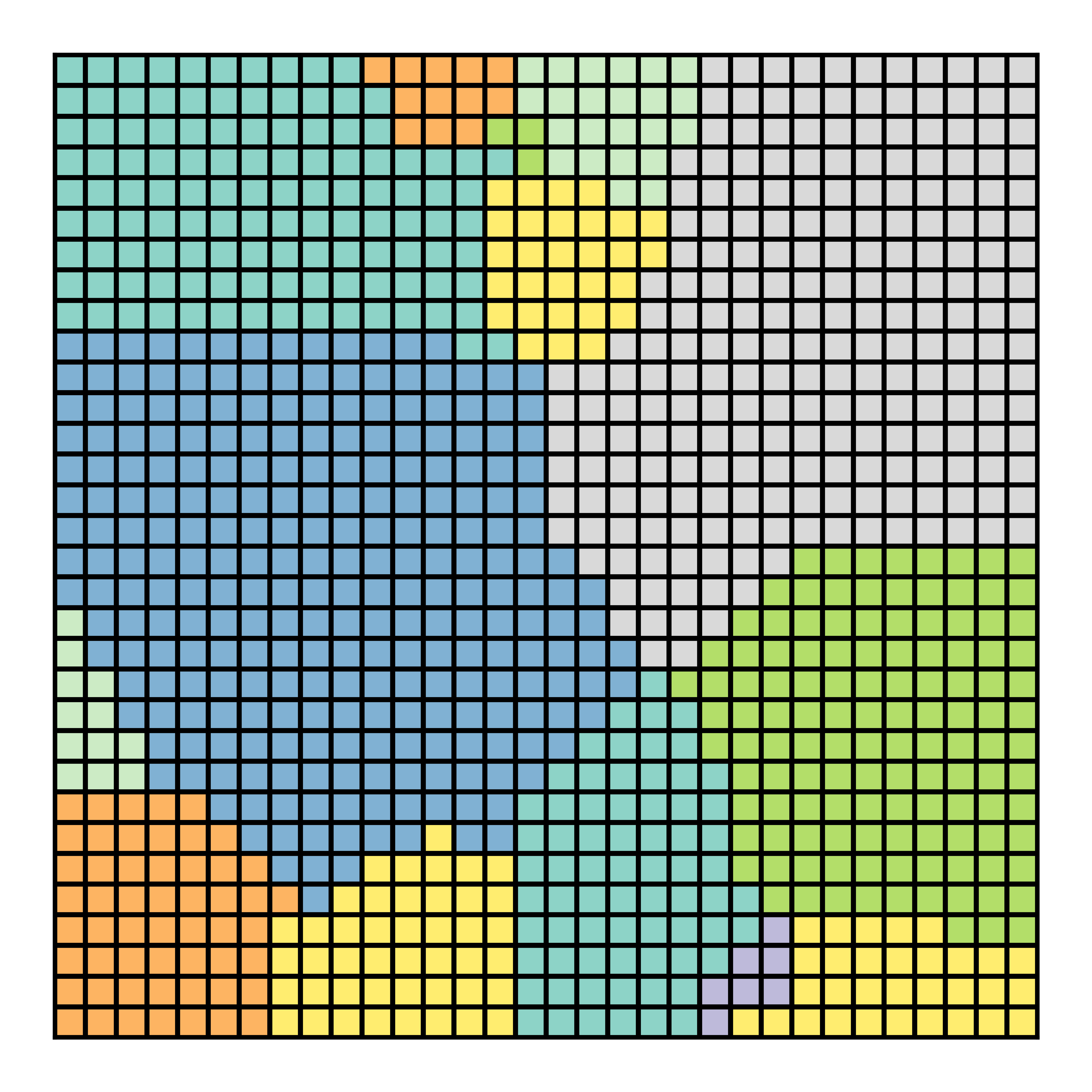}
\end{subfigure}

\begin{subfigure}[b]{0.30\textwidth}
\centering
\includegraphics[width=1.00\textwidth]{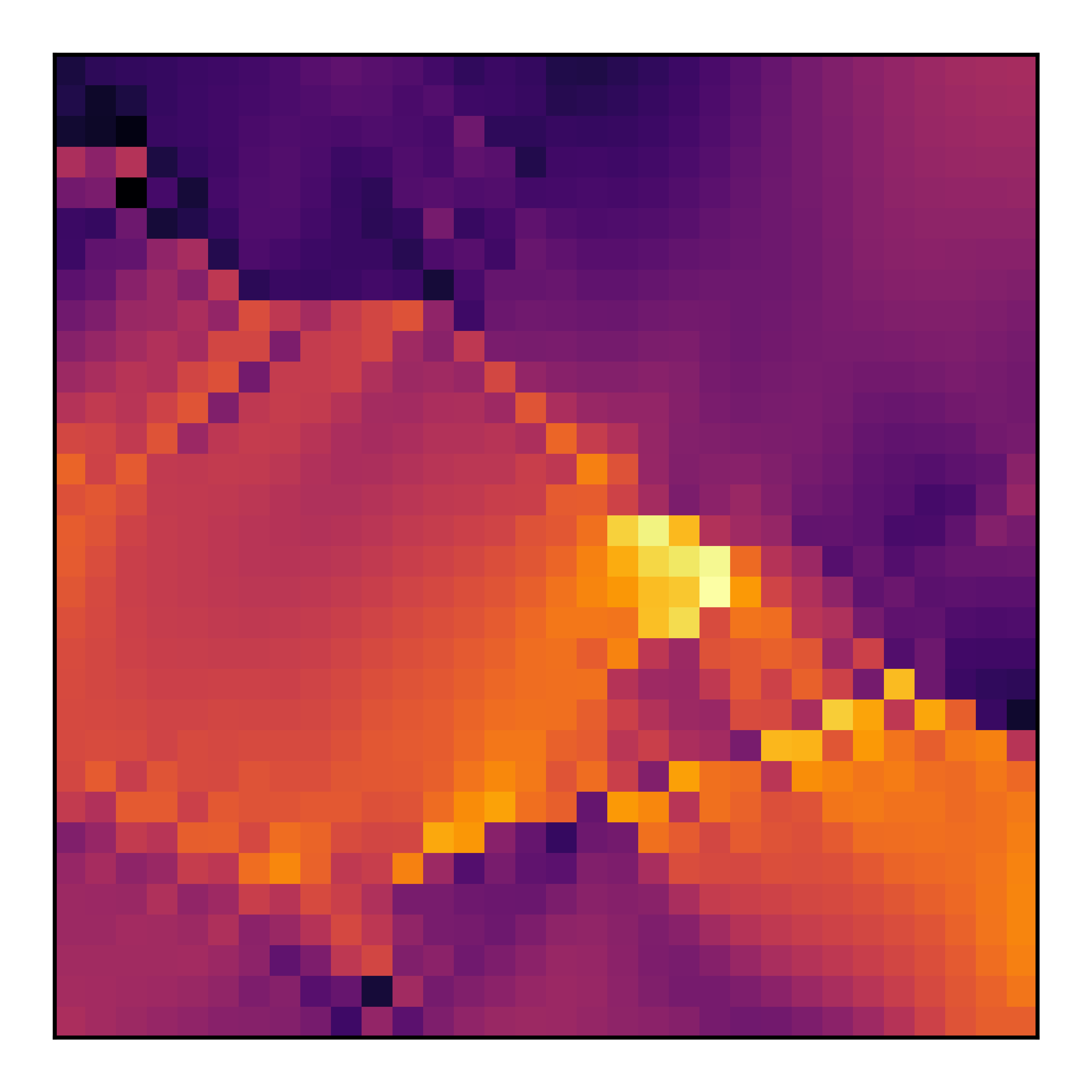}
\end{subfigure}
\begin{subfigure}[b]{0.30\textwidth}
\centering
\includegraphics[width=1.00\textwidth]{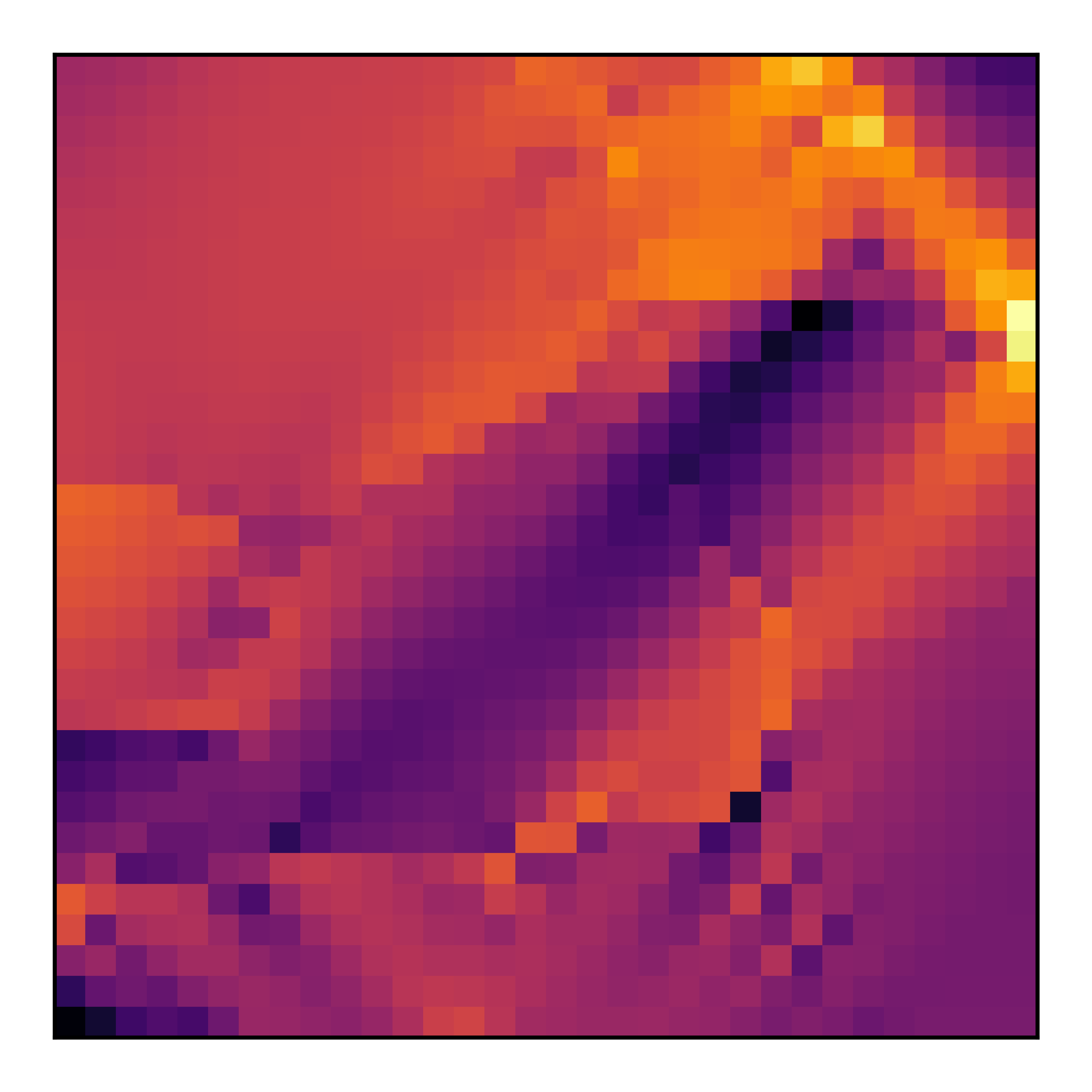}
\end{subfigure}
\begin{subfigure}[b]{0.30\textwidth}
\centering
\includegraphics[width=1.00\textwidth]{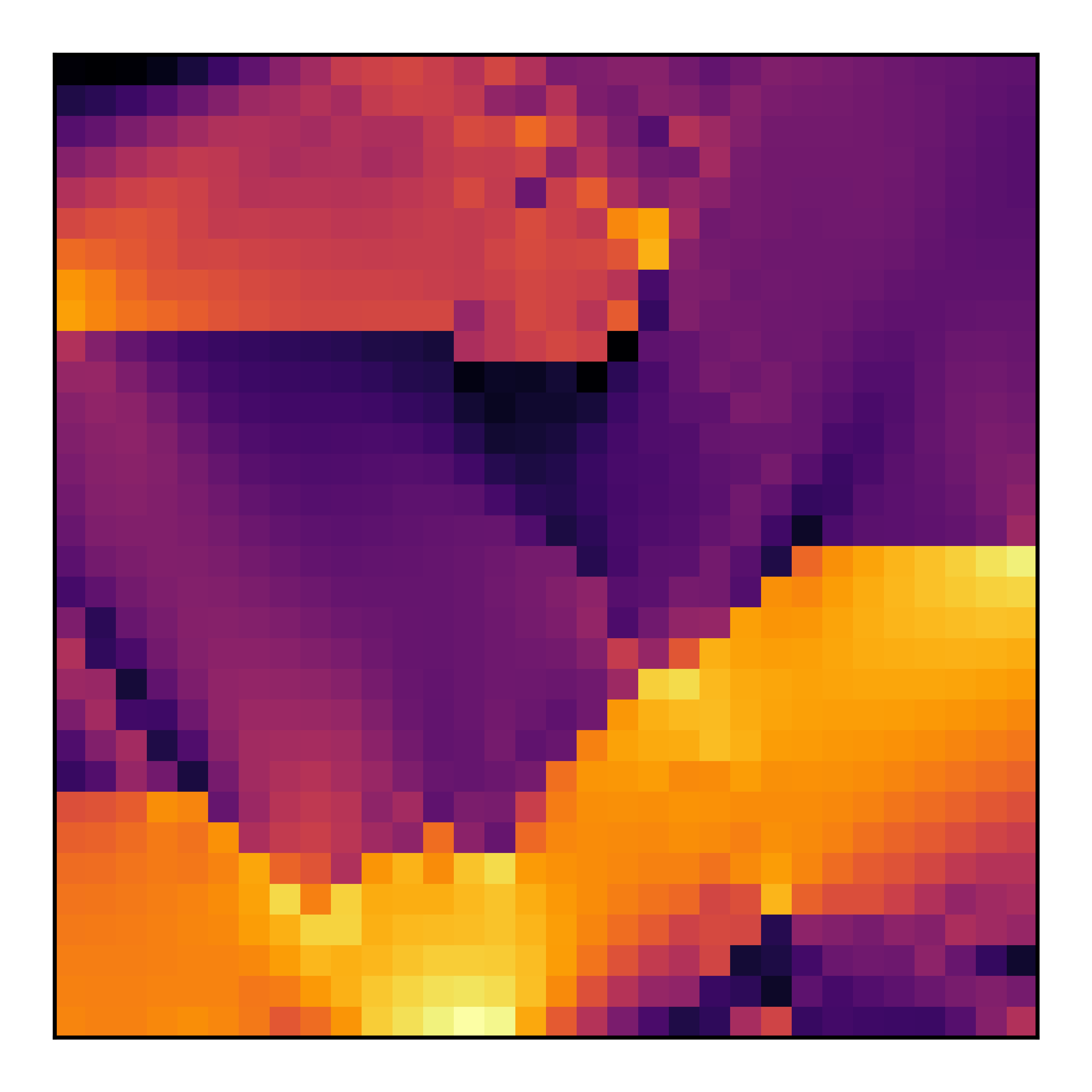}
\end{subfigure}
\caption{Crystal plasticity. Upper panel: realizations of the orientation field $\phi$. Lower panel: corresponding $\stress_{11}$ field at maximum applied tensile ($\strain_{11}$) strain.
}
\label{fig:cp_realizations}
\end{figure}

\begin{figure}[htb!]
\centering
\includegraphics[width=0.6\textwidth]{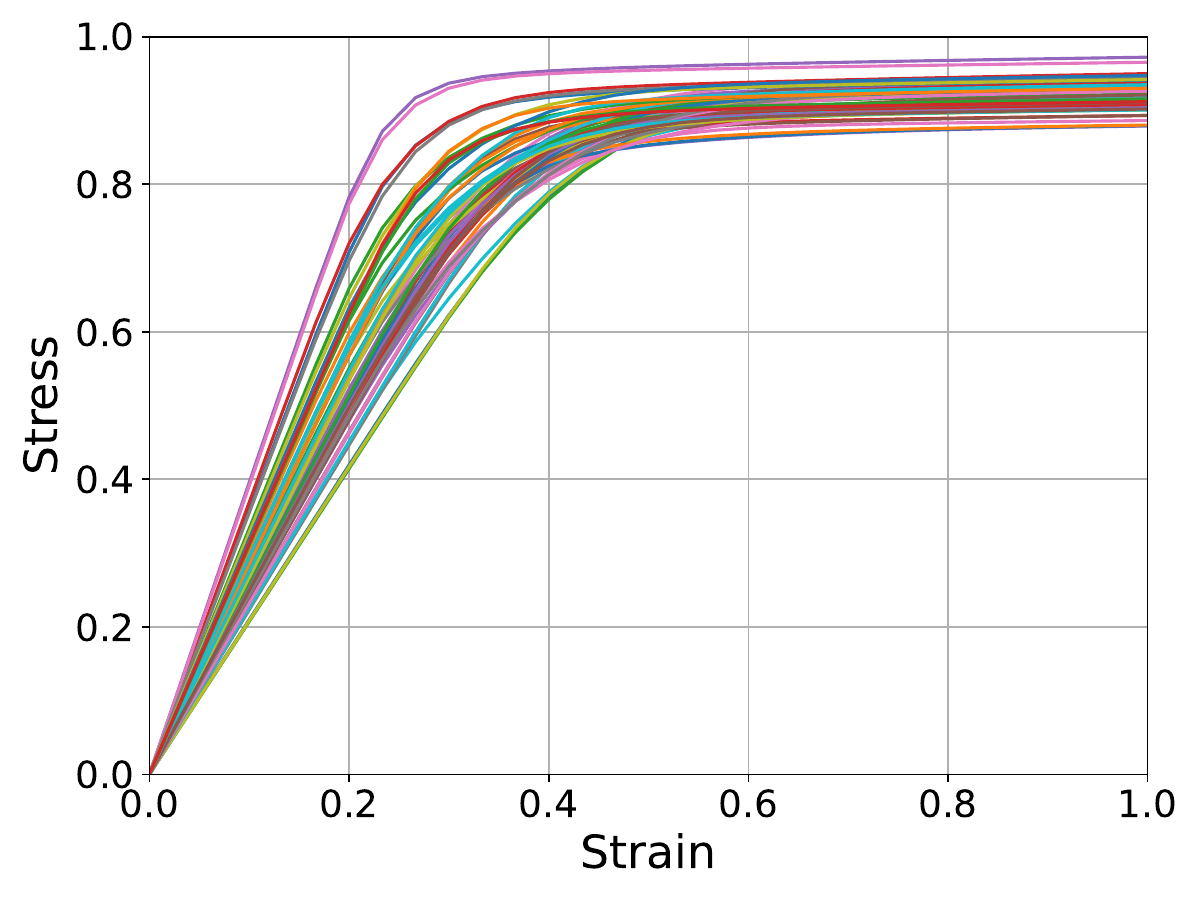}
\caption{Crystal plasticity: normalized stress response curves $\bar{\stress}(t)$ for different configurational realizations ($\phib(\Xb)$).
}
\label{fig:cp_responses}
\end{figure}

\fref{fig:cp_realizations}a shows 3 realizations of the orientation field $\phib$ on a 32$\times$32 grid which differ in the grain topology and the orientation field defining the grain structure.
For each sample the balance of linear momentum
\begin{equation}
\div \stress = \mathbf{0}
\end{equation}
is solved for the displacement field using the finite element method, subject to tension at a strain rate of $1/s$ effected by minimal boundary conditions  on the domain $\Omega_\square$, i.e., two parallel boundaries and one corner are constrained.
\fref{fig:cp_realizations}b illustrates the inhomogeneities in the stress response (at 0.3\% strain in tension), which are particularly marked at grain boundaries.
The overall response of samples reflects the anisotropy of each crystal which share boundaries and, hence, have kinematic constraints on their deformation.
The resulting mean stress $\bar{\stress}(t)$ response curves
\begin{equation}
\bar{\stress}(t) = \frac{1}{V} \int_\Omega \stress \, \mathrm{d}^3X
\end{equation}
are the final QoIs.
They vary due to the particular texture $\phi(\Xb)$ of the realization, where $V$ is the volume of $\Omega_\square$.
In fact, the different crystal textures evoke different effective elastic slopes, yield points, and (post-peak) flow stresses as can be seen in \fref{fig:cp_responses}.

\subsubsection{Neural network architecture} \label{sec:gru}

\begin{figure}[htb!]
\centering
\includegraphics[width=0.9\textwidth]{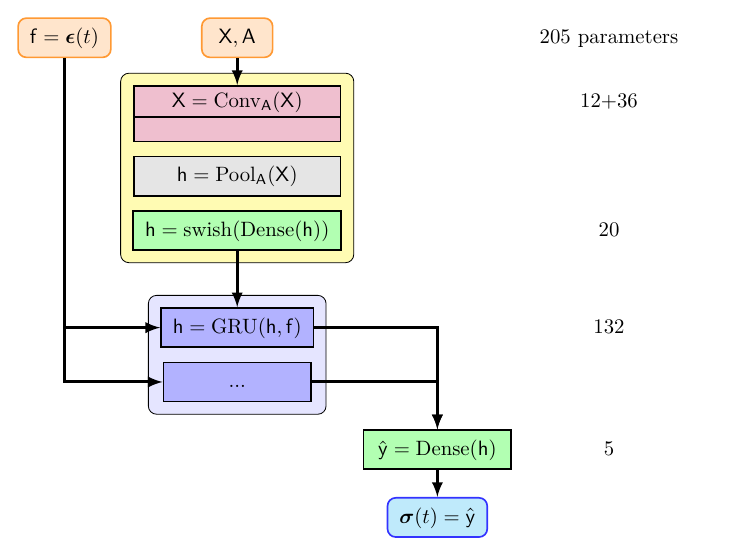}
\caption{GCNN-GRU model of the crystal plasticity data, $N_w = 205$.
The 2 convolutional layers have 4 filters each and \emph{swish} activation.
The GRU has $\tanh$ activation and is depicted as unrolled with a layer per time-step.
}
\label{fig:cp_arch}
\end{figure}

The general NN architecture for this data has been published previously~\cite{frankel2019oligocrystals,frankel2022mesh} and is depicted in \fref{fig:cp_arch}.
The physical inputs are the orientation field $\phi(\Xb)$ and the applied strain $\timevector=\bar{\strain}(t)$; the output is the mean stress over time $\outputvector=\bar{\stress}(t)$.
The orientation field $\phi(\Xb)$ is discretized on a mesh.
The discrete values of the texture $\phi$ and the associated cell volume fractions $\nu$ comprise the nodal data $\inputvector = [ \phi, \nu ]$ for the graph associated with discretization mesh.
The nodal input data $\inputvector$ and the graph adjacency matrix $\As$ are fed into a convolutional stack.
The GCNN layers have the form of the Kipf and Welling GCN \cite{kipf2016semi} with an independent self weight.
The result of this convolutional stack is pooled with global averaging to produce hidden features $\hiddenvector_0$ that describe the factors that are salient to the prediction of the mean stress.
One feature per filter is produced by the convolutional encoder.
These features depend only on the initial microstructure as described by $\phi(\Xb)$ and, hence, are time-independent.
These hidden features $\hiddenvector_0$ and the time dependent input $\timevector=\bar{\strain}(t)$ are concatenated and then fed into a GRU RNN \cite{cho2014learning} with the applied strain acting as $\timevector$.
The output of this layer is fed to a linear dense decoder to produce $\hat{\outputvector} = \bar{\stress}(t)$.
The entire model has a total of $N_w=205$ parameters.
\fref{fig:cp_arch} provides a schematic of the overall NN model with further details.

We will also employ a larger version of the model shown in \fref{fig:cp_arch} to demonstrate the performance of SVGD and pSVGD on a problem that is intractable to HMC.
This GCNN-RNN model operates on three-dimensional data using 2 convolutional layers with 16 filters, 1 post pooling dense layer with \emph{swish} activation, and 1 dense output layer.
It has 2609 parameters total, with $112+528$ in the convolutions, $272+17$ in the dense layers, and  1680 in the GRU.

\subsection{Diffusion and generation in polycrystals} \label{sec:diffusion}

Gas generation and diffusion in granular materials presents an engineering challenge in applications such as uranium fuel
\cite{aagesen2020phase,kim2022modeling}.
Predictive modeling of out-gassing can avert catastrophic failures in nuclear power infrastructure.
Here we reduce the learning problem to diffusion on graphs with a complex generation term.

\subsubsection{Physical system} \label{sec:gb_diffusion}
\begin{figure}[htb!]
\centering
\begin{subfigure}[b]{0.30\textwidth}
\centering
\includegraphics[width=1.00\textwidth]{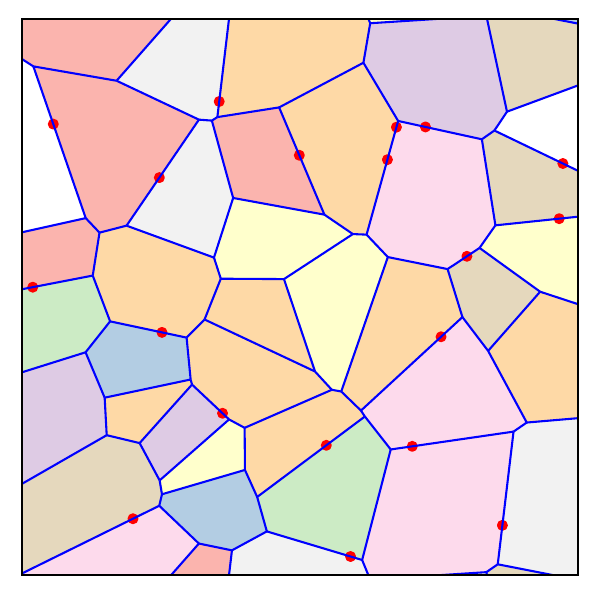}
\end{subfigure}
\begin{subfigure}[b]{0.30\textwidth}
\centering
\includegraphics[width=1.00\textwidth]{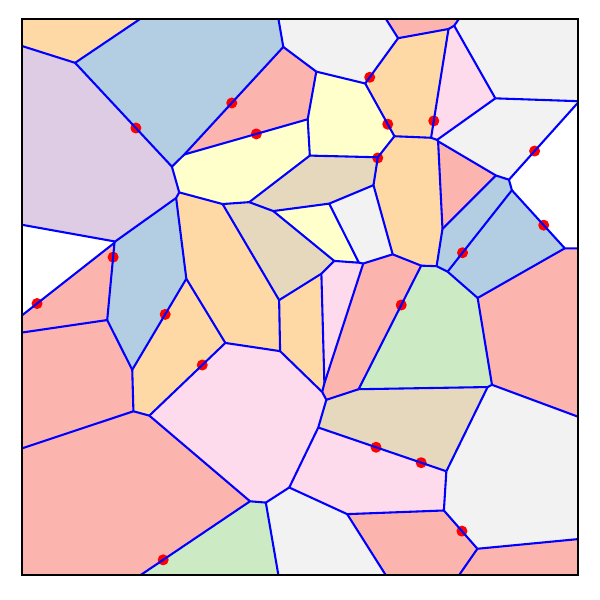}
\end{subfigure}
\begin{subfigure}[b]{0.30\textwidth}
\centering
\includegraphics[width=1.00\textwidth]{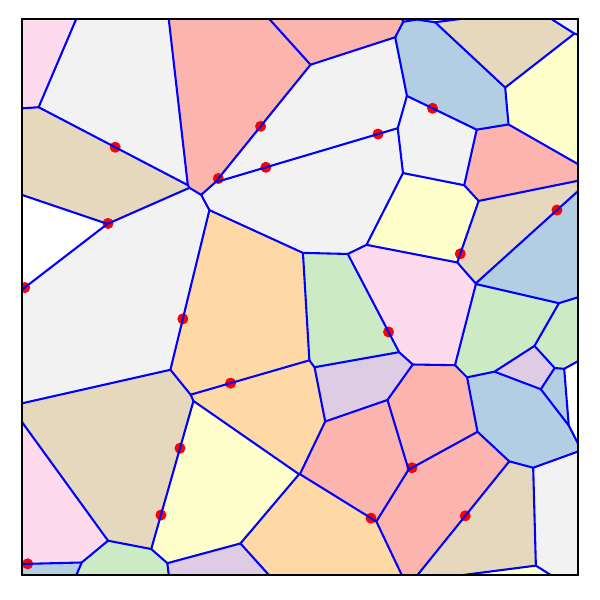}
\end{subfigure}

\begin{subfigure}[b]{0.30\textwidth}
\centering
\includegraphics[width=1.00\textwidth]{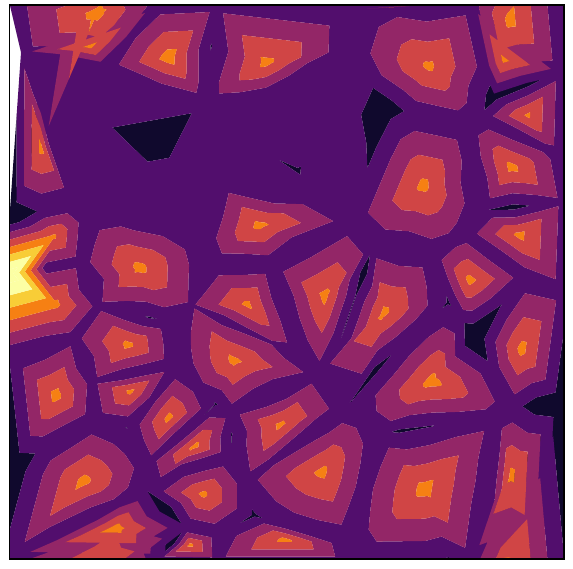}
\end{subfigure}
\begin{subfigure}[b]{0.30\textwidth}
\centering
\includegraphics[width=1.00\textwidth]{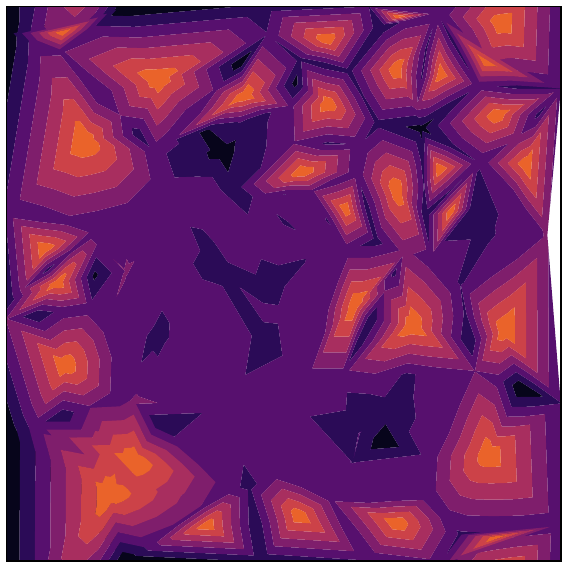}
\end{subfigure}
\begin{subfigure}[b]{0.30\textwidth}
\centering
\includegraphics[width=1.00\textwidth]{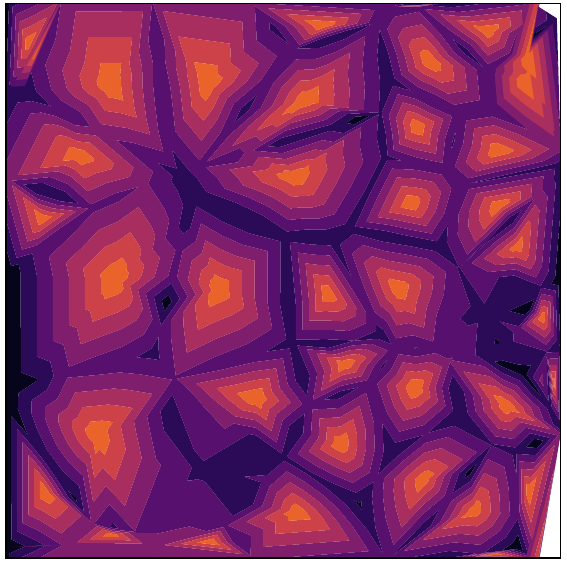}
\end{subfigure}
\caption{Gas release in polycrystalline systems.
Upper panel: initial state, red dots represent bubbles in voids.
Lower panel: gas concentration at $t=0.0005$.
}
\label{fig:gas_realizations}
\end{figure}

Similar to the crystal plasticity exemplar, the domain $\Omega_\square$ consists of a square with convex subregions $\Omega_K$ which represent grains; see \fref{fig:gas_realizations}.
Gas is generated in grains and slowly diffuses to the intergranular boundaries.
The diffusion along the more disordered, looser atomically packed boundaries is relatively fast.
This creates a intergranular network.

\begin{figure}[htb!]
\centering
\includegraphics[width=0.6\textwidth]{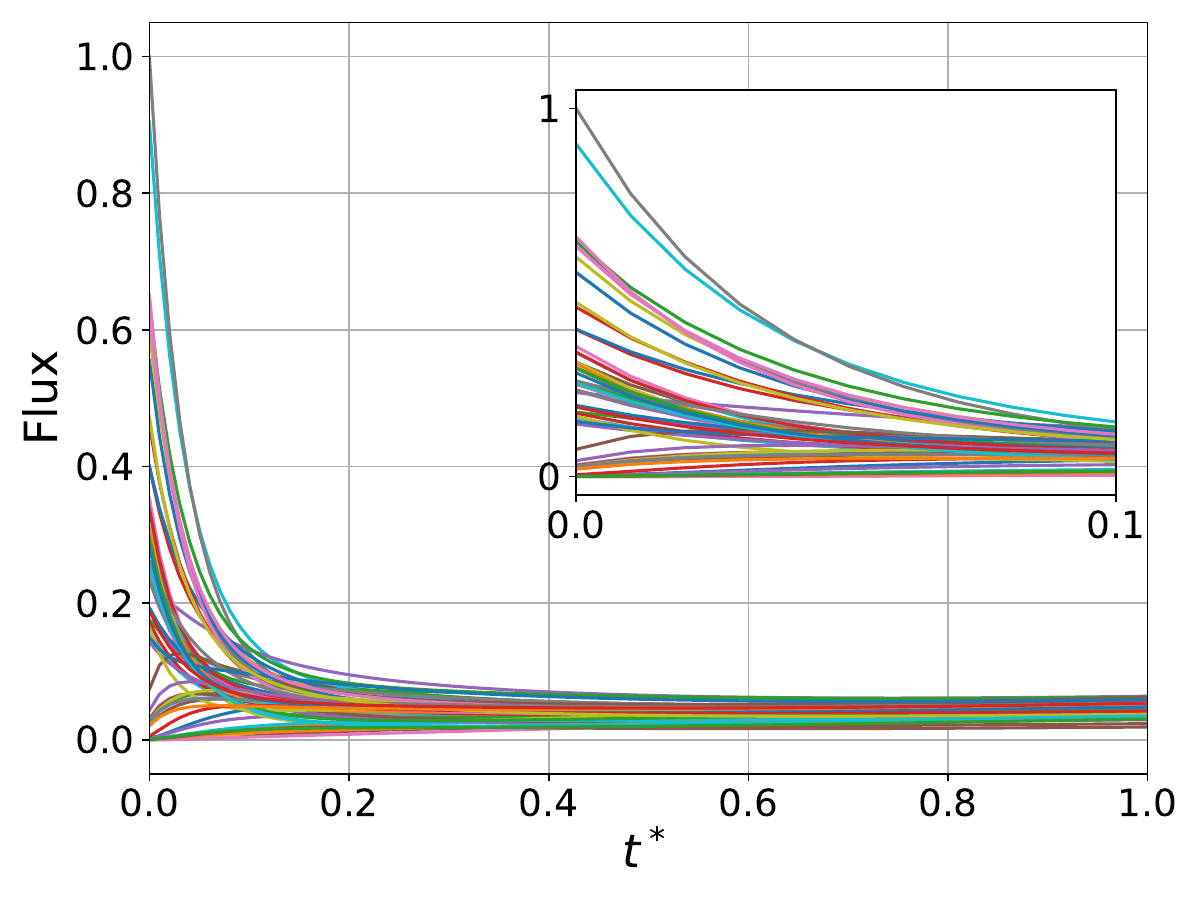}
\caption{Gas release: normalized flux response curves for different configurational realizations.
}
\label{fig:gas_responses}
\end{figure}

The coupled reaction-diffusion equations describing the concentration $\cs$ are
\begin{alignat}{2}
\dot{\cs} &=  \div( \kappa_b \grad \cs ) + \rho & \text{in} \ \Omega_K \\
\dot{\cs} &=  \div( \kappa_s \grad \cs ) & \text{on} \ \partial \Omega_K
\end{alignat}
where $\kappa_b$ is the bulk diffusivity, $\rho$ is the gas generation and $\kappa_s$ is the boundary diffusivity.
\tref{tab:gas_parameters} gives non-dimensional values for the parameters.
With reference to \fref{fig:gas_realizations}, the boundary conditions are: the top and bottom are periodic, the  left is open to out-gas, and the right boundary is closed.
Initially there are gas concentrations at voids in the material as depicted in \fref{fig:gas_realizations}a.
\fref{fig:gas_realizations}b shows the state of the extended system long after the initial conditions, and  \fref{fig:gas_responses} shows the variety of flux response curves obtained via a finite volume method.

\begin{table}[htb!]
\centering
\begin{tabular}{|l|cr|}
\hline
Intragrain gas generation      & $\rho$ &  700.0 \\
\hline
Intragrain gas diffusivity     & $\kappa_b$ &    1.0 \\
Grain boundary gas diffusivity & $\kappa_s$ & 1000.0 \\
\hline
\end{tabular}
\caption{Gas diffusion parameters representative of slow bulk diffusion and fast boundary diffusion, as in gas release from uranium fuel.
All are made dimensionless using $\kappa_b$ and the fact that the samples where in unit squares.
The maximum time is set to $0.0005$.}
\label{tab:gas_parameters}
\end{table}

\subsubsection{Neural network architecture} \label{sec:node}

Given the relative time-scales of the bulk versus the boundary diffusion processes, we choose to reduce the problem to a model on the grain boundary network $\cup_K \partial \Omega_K$:
\begin{equation}
\dot{\cs}  = \divg \cs + \rs(\cs)
\end{equation}
where  $\divg$ is the divergence operator on a graph and the generation term is intended to represent the complex interaction of the boundary network with the gas-generating grains.
After discretization, this system reduces to a system of ODEs with a linear-plus-nonlinear right-hand side:
\begin{equation} \label{eq:discrete_diffusion}
\dot{\cs} = \Ks \cs  + \Rs(\cs,\Xs) \ .
\end{equation}
where $\cs$ is the concentration on boundary network.
Since the concentration field $\cs$ in the grains is hidden from the network, we augment the concentration state with additional degrees of freedom to form the initial state $\Xs$ in the style of an augmented NODE \cite{chen2018neural}.
The augmented state is initialized with the nodal concentrations $\cs$ at time $t=0$, tributary areas $\nu$ (for bulk interactions), tributary lengths $\ell$ (for surface interactions), and zeros for additional states such that $\Xs = [\cs_0,\nu,\ell,0,...]$ for every graph node, this information becomes the initial hidden state $\hiddenvector$.
Since there are no external influences on the system $\timevector$ is simply a sequence of times.

The decomposition in \eref{eq:discrete_diffusion} enables a physics informed treatment where we assume the form of the diffusion is defined by the graph describing the network and the generation term $\Rs$ is completely data informed.
The linear part of the right-hand side representing in-network diffusion is simply
\begin{equation}
\Ks \cs = \operatorname{ReLU}(\kappa) \Ls \cs \ ,
\end{equation}
where $\Ls = \operatorname{rowsum}(\As) - \As$ is the graph Laplacian, $\As$ is the graph adjacency, and $\kappa$ is a non-negative trainable parameter.
The nonlinear term $\Rs$ representing gas generation is modeled by graph convolutions so that the overall system is notionally
\begin{equation}
\dot{\cs} = \underbrace{ \operatorname{\kappa} \Ls}_{\Ks} \cs + \underbrace{\Conv \Xs }_{\Rs(\cs,\Xs)}
\end{equation}
where $\Conv$ is a stack of convolutions and activations.
The parameters of the convolutions $\Conv$ and $\kappa$ are the learnable aspects of this hybrid NODE.
Boundary conditions are imposed with penalization in $\Ls$ and by projection with $\Rs$.
The QoI $\outputvector$ is flux of gas at the open boundary over time, which is obtained by selective pooling of $\hiddenvector$.

The form of \eref{eq:discrete_diffusion} suggests a classical fractional-step/operator-split \cite{strang1968construction,chorin1968numerical,martyna1996explicit} time integration.
For the linear part we employ an exponential integrator.
For the nonlinear part we employ a forward Euler update, although backwards Euler, midpoint, etc. can also been utilized.
The complete time integration scheme for the concentrations $\cs$ and hidden variables is:
\begin{eqnarray}
\cs_{n+1/2}  &=& \exp(1/2 \dt \, \Ks) \ \cs_n  \nonumber \\
\hiddenvector_{n+1}    &=& \hiddenvector_n + \dt \, \Rs(\hiddenvector_n) \\
\cs_{n+1}    &=& \exp(1/2 \dt \, \Ks) \ \cs_{n+1/2} \nonumber
\end{eqnarray}
with the hidden state vector $\hiddenvector$ subsuming the (hidden) concentration state $\hiddenvector = [ \cs, \ldots ]$, and with slight abuse of notation $\Rs(\hiddenvector)$ is the learnable function emulating $\Rs(\cs,\Xs)$.

\begin{figure}[htb!]
\centering
\includegraphics[width=0.9\textwidth]{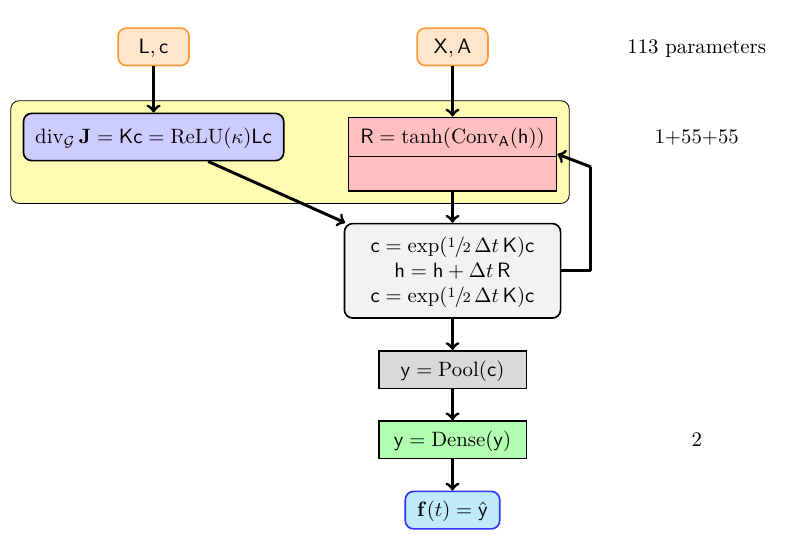}
\caption{Hybrid NODE-GCNN architecture for gas release.
Each convolutional layer has 5 filters which is also the size of the augmented state $\Xs$.
}
\label{fig:gcnn_node}
\end{figure}

The hybrid NODE model is depicted in \fref{fig:gcnn_node}, which also contains additional details of the architecture such as depth of the convolutional stack, activations, and number of filters.
Note that the time-scaling of the NODE allows for training strategies such as training to an increasing sequence of history or training to a coarse-then-fine version of the history to ease training to the full history.

\subsection{Sampling and training}

In this section we describe implementation details of our methods and hyperparameter choices.

The MAP, which are needed for initializing the particles in SVGD and pSVGD, are found using the Adam optimizer \cite{kingma2014adam} with early stopping.
In all numerical examples, we employ a prior to be $\prior(\weight)\sim \Nc(\mathbf{0},\sigma_0^2 \Ib)$ with $\sigma_0 = 1$.
The NN parameter prior's standard deviation of $\sigma=1$ is sufficiently large to be uninformative compared to the range $\mathcal{O}(10^{-2})$ to $\mathcal{O}(10^{-1})$ eventually explored by the NN parameters.
The likelihood follows an observation model  described in \sref{ss:Bayesian}:
$\outputvector_{i,n} = \outputvectorP_{i,n}(\inputvector_i,\timevector_i,\weight) + \epsilon_{i,n}$ with indepedent Gaussian noise $\epsilon_{i,n} \sim \mathcal{N}(0,\sigma_{\epsilon}^2)$.
The data noise is set to $\sigma_{\epsilon} = 0.05$ for the crystal plasticity data and  $\sigma_{\epsilon} = 0.02$  for the gas flux data.
All values of $\inputvector, \timevector, \outputvector$ are normalized to the range $[0,1]$ using their corresponding minimum and maximum instances from the training dataset.

To boost efficiency of HMC in terms of computational time and memory when handling the larger datasets, gradient evaluation of the likelihood, \eref{eq:HMC_potential}, is approximated by using a random subset of the full training dataset and proportionally scaled up by the number of data points. This subset is  resampled at every evaluation.
This procedure can be seen as a form of pseudo-marginal subsampling~\cite{dang2019hamiltonian}.

The SVGD initializes $N_p$ particles at the MAP plus some jitter to make them distinct.
We adopt a weighted RBF kernel
\begin{equation}
\kernel(\weight,\weight') = \exp\left(-\frac{1}{h}\| \weight - \weight' \|_{\tilde{\Hs}}^2\right) ,
\end{equation}
where $\tilde{\Hs}$ is an approximate Hessian matrix and $h$ is the bandwidth.
We select an adaptive bandwidth $h = \bar{d}^2/ \log N_p$ where $\bar{d}$ is the median pair-wise distance between particles.
pSVGD use the same initialization and employs a tolerance on the eigenvalues of the Hessian to construct a subspace from its eigenvectors.
The Hessian-based reductions are done layer-wise to prevent disconnected networks.

\section{Results} \label{sec:results}

In this section, we first we use the two-dimensional (2D) NN models described in \sref{sec:exemplars} to compare SVGD with HMC, which we treat as the reference baseline.
We then use the larger three-dimensional (3D) NN models to explore the performance of pSVGD and comparing with  SVGD, for a problem size that is currently infeasible for HMC.

The exemplar NN models described in \sref{sec:exemplars} are similar in that they all predict the evolution of a global QoI based on an initial spatial configuration.
In terms of parameters, the RNN-based model of plasticity (\sref{sec:plasticity}) has many parameters in the GRU, while the NODE model of diffusion (\sref{sec:diffusion}) has relatively more parameters in the graph convolutions than in the time integration.
Furthermore, the diffusion model has a physics-informed plus learned discrepancy formulation with a component with known form, the diffusion, and the learned part, the generation, and hence is more constrained than the GCNN-RNN.
The GRU effectively has the same initial conditions, see \fref{fig:cp_responses}, while the NODE has a wide variety, see \fref{fig:gas_responses}.
For the GRU model of plasticity, the hardest feature to capture is the transition from elastic to fully plastic flow, while the NODE has the task of correlating the graph structures and bubble locations with the variety of initial conditions.
Both formulations can have stability issues predicting the evolving processes.

\subsection{Comparison of SVGD and HMC} \label{sec:comp_hmc}
In this section we compare the UQ results provided by SVGD and HMC for the two exemplars.
Since many NNs tend to be over-parameterized, we examine two cases of different training data sizes: (a) $N_D = \frac{1}{2} N_w$ with the number of samples being half the number of parameters, and (b) $N_D = 2 N_w$ with the number of samples being twice the number of parameters.
For SVGD we use  $N_p = 96$ particles, which is on par with the number of parameters.

\begin{figure}[htb!]
\centering
\includegraphics[width=0.4\textwidth,trim={6cm 15 5.5cm 15},clip]{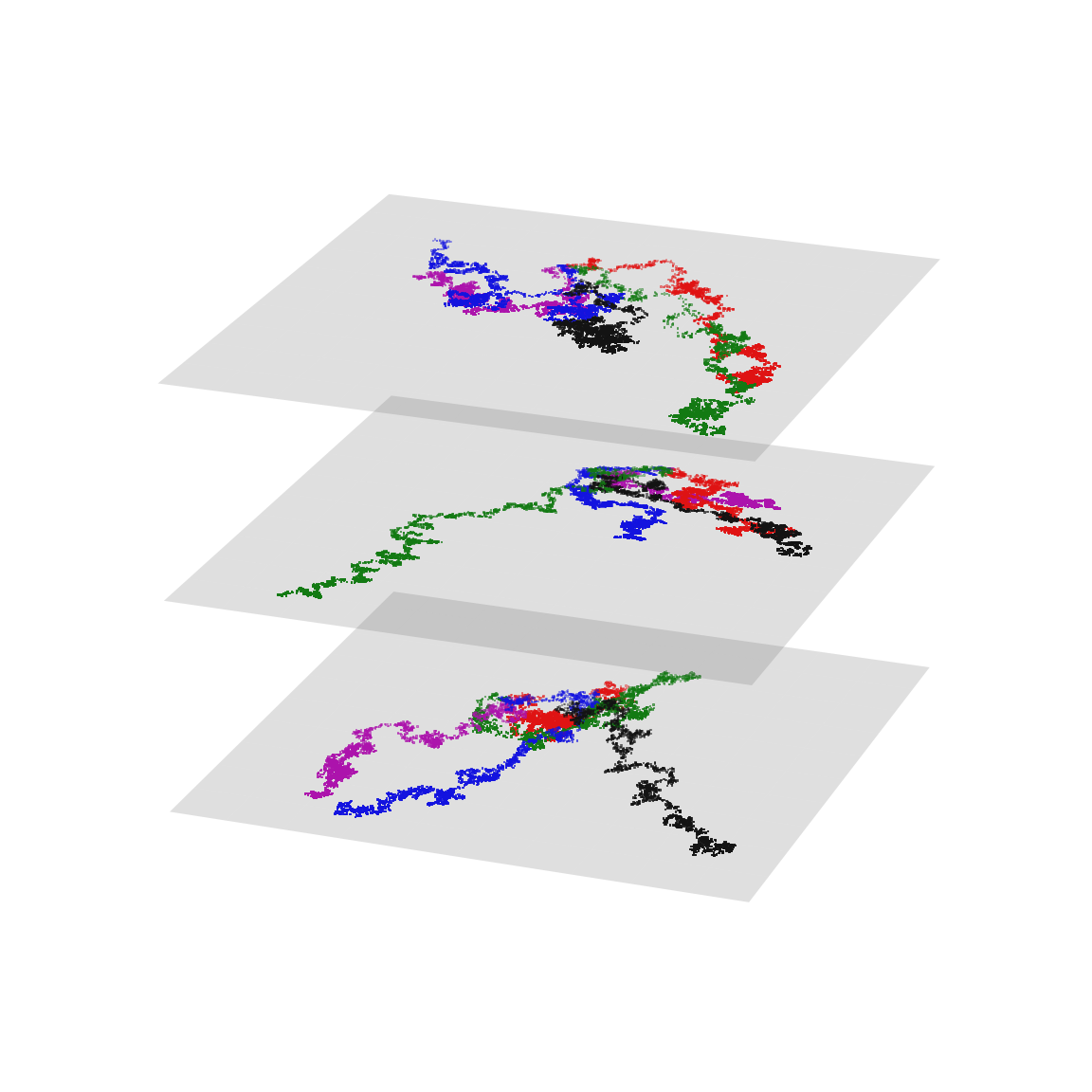}
\includegraphics[width=0.4\textwidth,trim={6cm 13 5.5cm 13},clip]{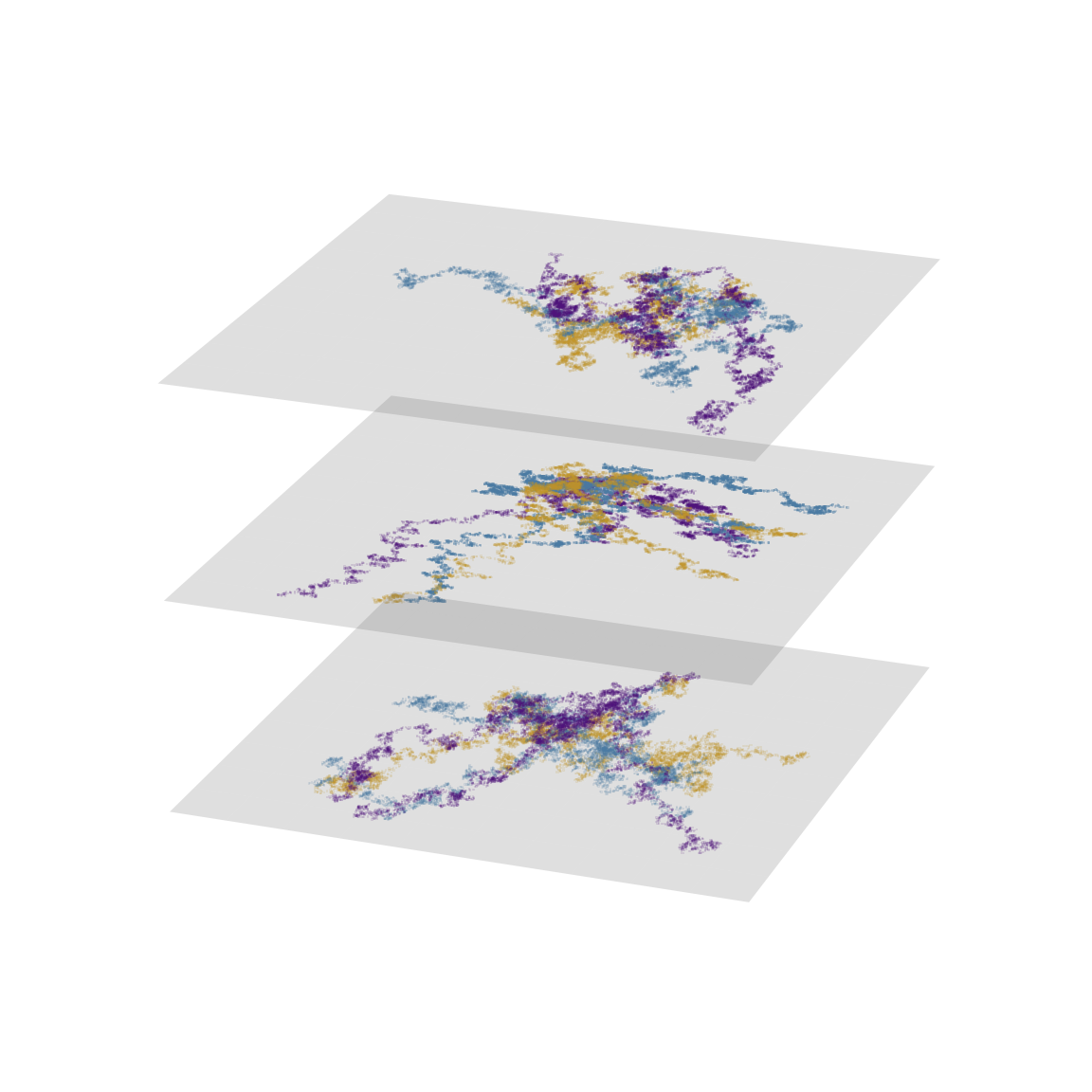}
\caption{HMC chain traces plotted in two parameter components for the GRU model showing the dynamics for several parameter pairs. Left frame shows chain samples that start at the same MAP value, with different random number generator seeds. The right frame shows samples obtained from chains initialized from different MAP values. Results were thinned out (shown every $10^3$ samples) for clarity.
}
\label{fig:map}
\end{figure}

\fref{fig:map} shows traces of HMC chains for the crystal plasticity exemplar problem modeled with a GCNN-GRU.
While each 2D slice individually would suggest the chains cover the same parameter region, it turns out that each set of results explore a separate region when inspecting chains across several slices of the parameter space.
Each individual chain, either starting from a common MAP value (in the left frame) or starting from multiple MAP values (in the right frame) exhibit a high degree of autocorrelation induced by the seemingly non-convex posterior distribution. Recall that these MAP values are found using gradient-based numerical optimization algorithms, and thus may correspond to different \emph{local} MAP values (i.e., from local minima).
Results for the NODE-based model are similar to the GRU-based results shown in \fref{fig:map}.
While the HMC stepsize and number of steps were adapted to obtain reasonable acceptance rates and stable NN models, further attempts to alleviate the degree of autocorrelation have not been successful.

\begin{figure}[htb!]
\centering
\includegraphics[width=0.3\textwidth,trim={2cm 3.5cm 0.8cm 5.2cm},clip]{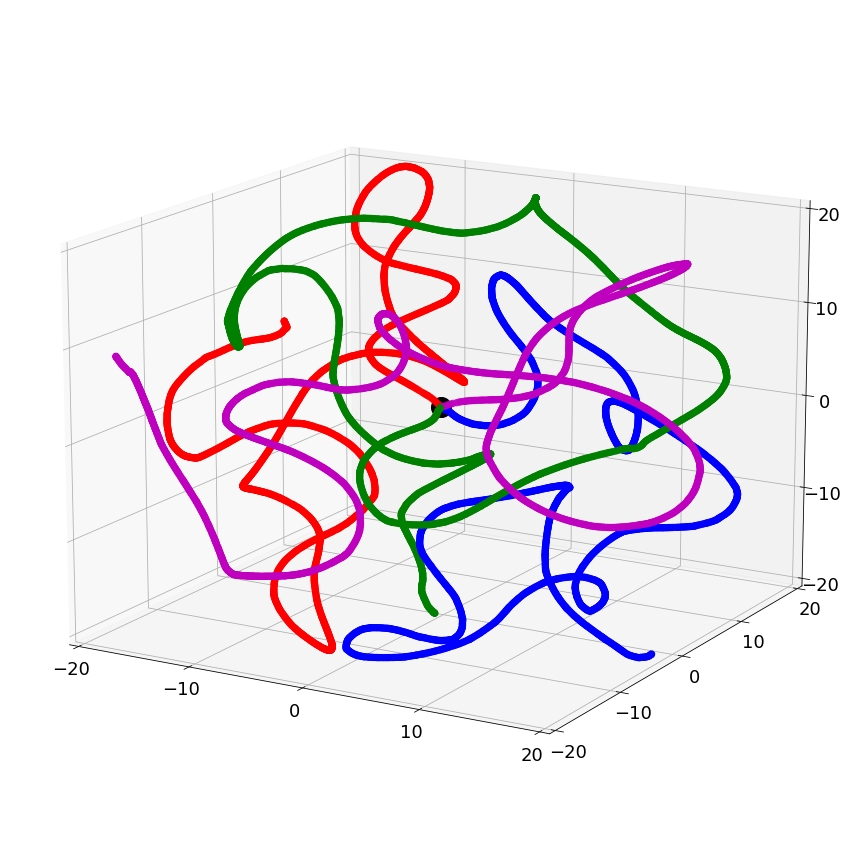}
\hspace{1cm}
\includegraphics[width=0.3\textwidth,trim={1cm 1.5cm 1cm 2.5cm},clip]{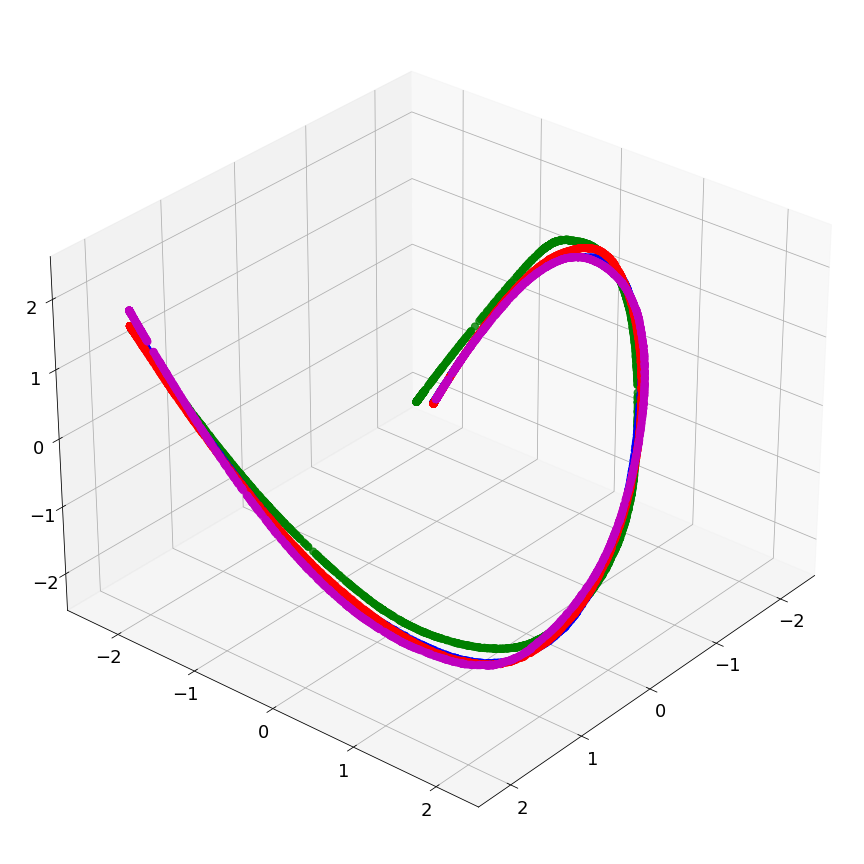}
\caption{GRU: Low-dimensional embeddings for the HMC chains corresponding to the case using $N_{d}=\frac{1}{2} N_{w}$. Results correspond to 4 chains with different random seeds, all starting from the same MAP value (location shown with black circle in the left frame). The left frame shows data processed with the T-SNE algorithm; the right frame shows results generated via spectral embedding.
}
\label{fig:rnn_submanifold}
\end{figure}

\fref{fig:rnn_submanifold} shows low-dimensional representations of the chains used to construct the 2D slices in \fref{fig:map}.
The left panel depicts 3D t-SNE representations~\cite{Maaten:2008} of chains starting from the same local MAP value, shown with a black circle.
This dimensionality reduction technique clusters high-dimensional samples onto a lower dimensional space by minimizing the Kullback-Liebler divergence between probabilities that define the degree of similarity between samples in the high- and the low-dimensional spaces, respectively.
The reduction to a 3D space, in the right panel, shows HMC paths that do not mix, consistent with the chain traces illustrated in \fref{fig:map}.
The right frames shows low-dimensional representations obtained via Laplacian eigenmaps~\cite{Belkin:2003} constructed using pair-wise distances between HMC samples.
This dimensionality reduction technique is similar to the the diffusion map framework employed in \sref{sec:illustration} for the canonical distribution samples.
The Laplacian eigenmap shows the components of the eigenvectors corresponding to the largest three eigenvalues of the Laplacian matrix.
The eigenvectors corresponding to the four chains exhibit a similar structure when analyzed independently which indicate that chain evolve along a narrow high posterior region but never return to the starting point or any region previously visited.

We next employ the distance correlation~\cite{szekely2007measuring,szekely2009brownian} to measure the degree of dependence between subsets of parameters in the GRU and NODE models.
By construction, distance correlation can be applied to random variables of different dimensionality.
\tref{tab:dist_corr_gru} and \tref{tab:dist_corr_node} give the distance correlation between the weight ensembles grouped by layer.
The entries in these tables are ranges covering simulations starting from several MAP values.
For GRU, all layer parameters are highly dependent, with the output having a slight increasing trend from input convolution (\emph{GCNN1}) to the pre-output dense layer.
The correlations for NODE on the other hand, show that the graph convolutions are strongly dependent (the bold entries in the table) and work together to represent the complex generation behavior.
These layers are only loosely correlated with the single parameter (\emph{Scale}) used in the diffusion component and with the output layer.
The NODE also shows increasing dependence of the layers from input to output with the output layer, albeit much weaker compared to the GRU model.
\begin{table}[htb!]
\centering
\begin{tabular}{l|cccc||cccc}
&  GCNN1      & GCNN2 & Dense & GRU
&  GCNN1      & GCNN2 & Dense & GRU         \\ \hline
GCNN2  &  0.90--0.94 &       &       &
&  0.92--0.93 &       &       &   \\
Dense  &  0.91--0.92 & 0.93--0.96 &  &
&  0.91--0.94 & 0.93--0.96 &  &  \\
GRU    &  0.92--0.94 & 0.97--0.98 & 0.95--0.98 &
&  0.91--0.93 & 0.96--0.97 & 0.96--0.97 &   \\
Output &  0.72--0.73 & 0.78--0.83 & 0.81--0.85 &  0.81--0.85
&  0.58--0.83 & 0.75--0.85 & 0.75--0.89 &  0.78--0.89
\end{tabular}
\caption{Distance correlation values for the HMC-based inference using the GRU model and $N_D=105$ (left) and $N_D=420$ (right) data samples.}
\label{tab:dist_corr_gru}
\end{table}

\begin{table}[htb!]
\centering
\begin{tabular}{l|ccc||ccc}
& Scale & GCNN1 & GCNN2 & Scale & GCNN1 & GCNN2 \\ \hline
GCNN1  &  0.54--0.71 & - & - &  0.55--0.79 & - & - \\
GCNN2  &  0.50--0.69 & {\bf 0.97--0.98} & - &  0.55--0.69 & {\bf 0.96--0.98} & -\\
Dense  &  0.15--0.24 & 0.22--0.47 & 0.23--0.45 &  0.14--0.30 & 0.20--0.25 & 0.21--0.27\\
\end{tabular}
\caption{Distance correlation values for the HMC-based inference using the NODE model and $N_D=60$ (left) and $N_D=240$ (right) data samples.}
\label{tab:dist_corr_node}
\end{table}

\begin{figure}[htb!]
\centering
\includegraphics[width=0.9\textwidth]{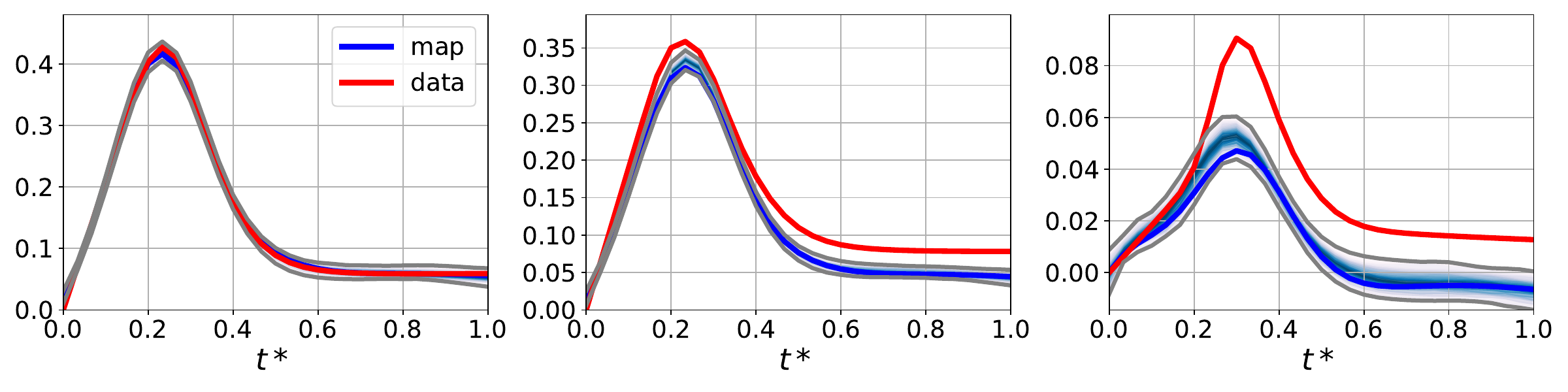}
\includegraphics[width=0.9\textwidth]{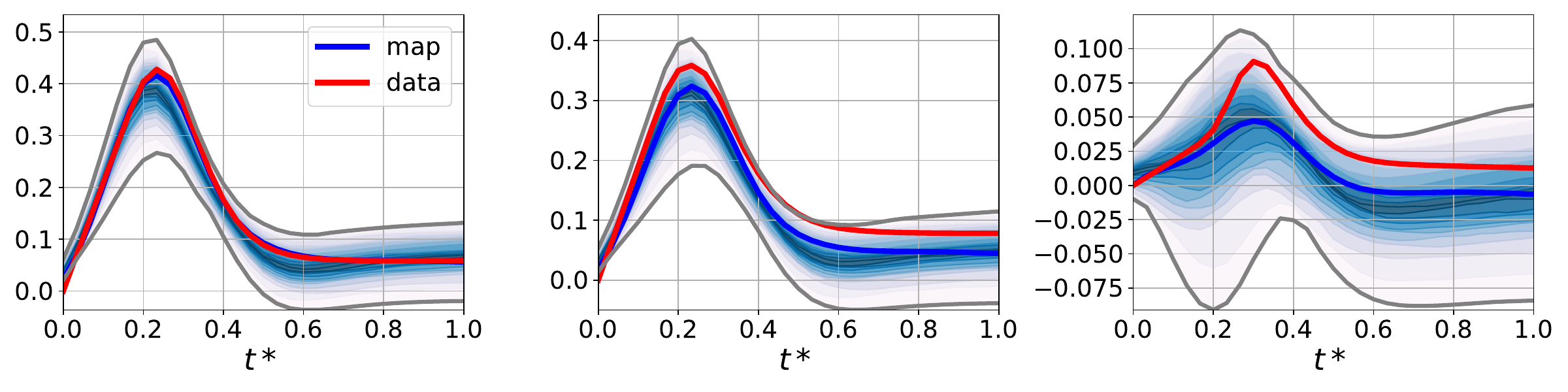}
\caption{RNN: Three cases corresponding to minimum (left), median (middle), maximum (right) MAP discrepancy for training with $N_D = \frac{1}{2} N_w$ data points. Each plot show the MAP prediction, posterior push-forward distribution, and the corresponding data sample. Top row shows HMC results and bottom row shows SVGD results.
Note the change in vertical scale from left panels to the right.
}
\label{fig:rnn_pf}
\end{figure}

\begin{figure}[htb!]
\centering
\includegraphics[width=0.9\textwidth]{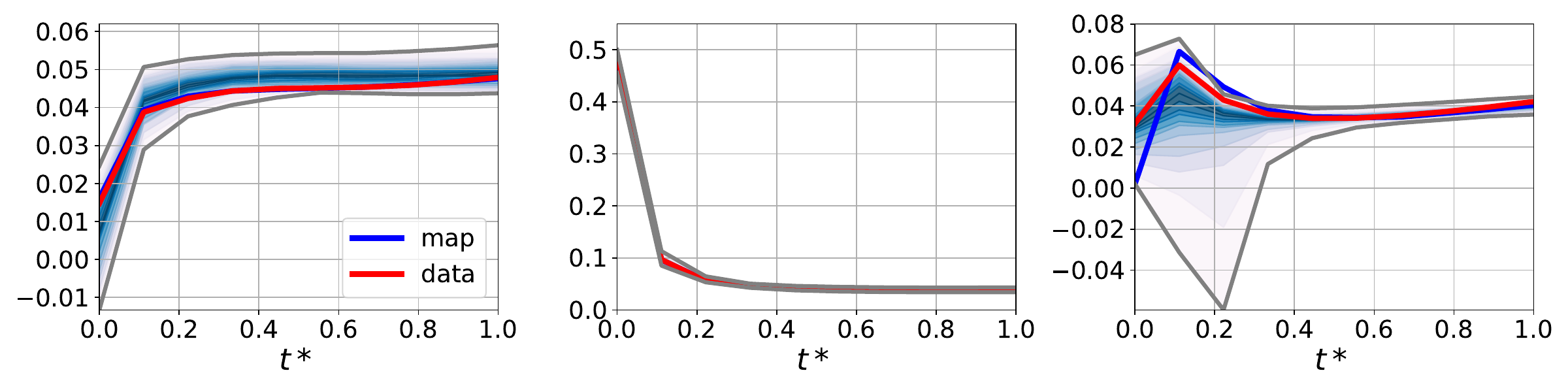}
\includegraphics[width=0.9\textwidth]{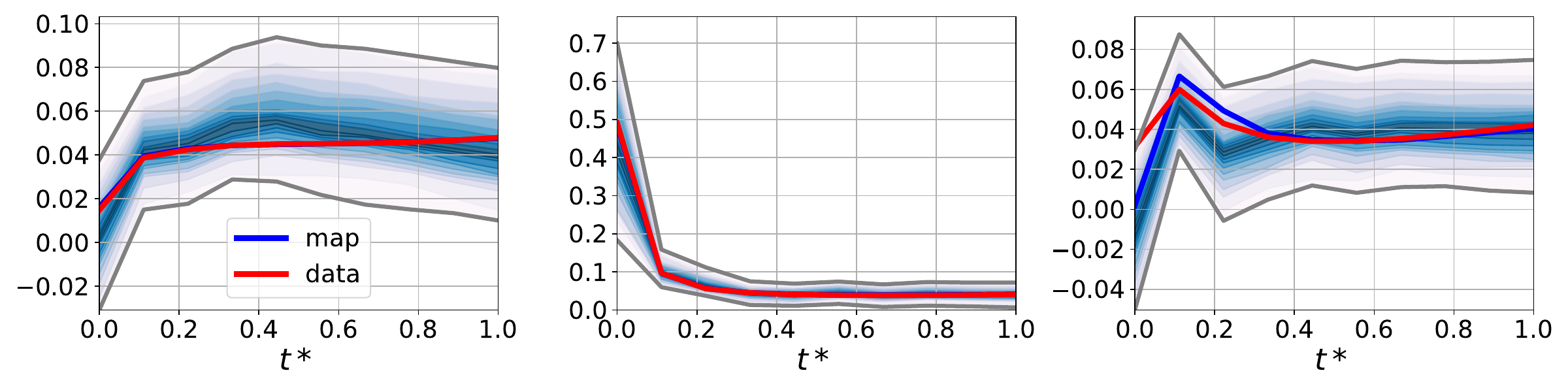}
\caption{NODE: Three cases corresponding to minimum (left), median (middle), maximum (right) MAP discrepancy for training with $N_D = \frac{1}{2} N_w$ data points. Each plot show the MAP prediction, posterior push-forward distribution, and the corresponding data sample. Top row shows HMC results and bottom row shows SVGD results.
Note the change in vertical scale from left panels to the right.
}
\label{fig:node_pf}
\end{figure}

\begin{figure}[htb!]
\centering
\includegraphics[width=0.8\textwidth]{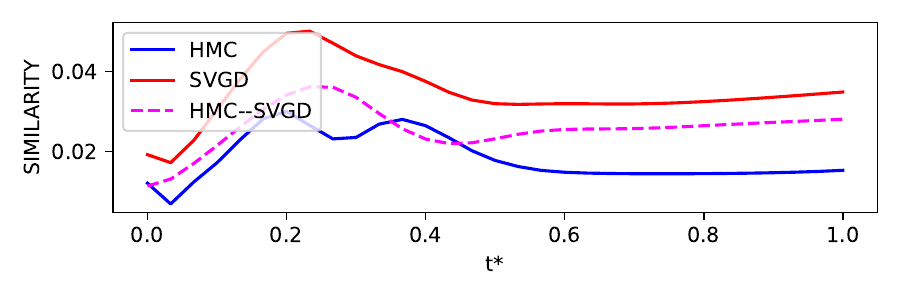}
\includegraphics[width=0.8\textwidth]{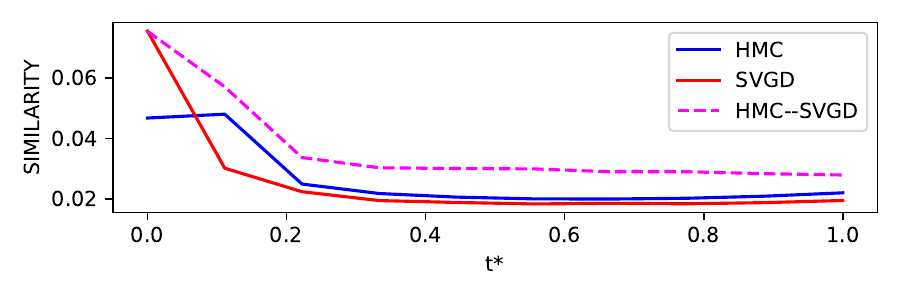}
\caption{
KSS similarity measures for (a) HMC versus observed data, (b) SVGD versus observed data, and (c) HMC versus SVGD; GRU (upper row) and NODE (lower row). A lower score indicates a greater similarity.}
\label{fig:distribution_dist}
\end{figure}

\fref{fig:rnn_pf} and  \fref{fig:node_pf} show the GCNN-RNN and GCNN-NODE model predictions of the QoI made at the MAP realization $\weight_{\text{MAP}}$, for three cases corresponding to minimum, median, maximum MAP discrepancy from the entire dataset.
We define the MAP discrepancy, for the $i$th data sample, as the root-mean square error (over all time steps in the time trace) between the MAP prediction compared to the corresponding data sample:
\begin{equation}
\| \outputvector_i - \NN(\inputvector_i,\timevector_i; \weight_\text{MAP}) \|_\text{RMSE} .
\end{equation}
The colormap indicates higher probability with darker shades near the median of the PF distribution.
The grey lines correspond to the $5\%$ and $95\%$ quantile levels.
The solid blue line indicating the MAP PF should generally overlap with the darker blue region.
Nevertheless, in this case, as the posterior distribution is highly irregular, different chain discover adjacent local minima, giving rise to the discrepancy between the initial condition for these HMC chains and the bulk of the subsequent samples.
The confidence bands of the PFs show some signature of uncertainty growing in time and being largest where there is the most variance in the data.
The confidence bands also show that  SVGD appears to  better cover the data; however, the HMC and SVGD predictions are more comparable when the entire dataset is taken into account, as shown in \fref{fig:distribution_dist}.

\fref{fig:distribution_dist} illustrates the similarity between (a) HMC versus observed data, (b) SVGD versus observed data, and (c) HMC versus SVGD results, over time.
We use the Kolmogorov-Smirnov statistic (KSS) to measure the similarity of two cumulative distribution functions (CDFs):
\begin{equation} \label{eq:kss}
\operatorname{KSS}(\CDF_1,\CDF_2) = \int_{-\infty}^\infty | \CDF_2(y) - \CDF_1(y) | \, \mathrm{d}y,
\end{equation}
where we approximate the CDFs with the empirical CDFs constructed with the posterior push-forward distributions using samples from the HMC chains and SVGD particles, respectively.
To facilitate the comparison of a posterior push-forward  with data, we use a Heaviside function centered at the data value $y$ as the data CDF.
These statistics also show the signatures of the variance of the output features through time.
Similarities between the predictions and the data are good throughout time but are more dissimilar (higher KSS) in the highly variable, hard to predict regimes.
The highly variable plastic transition in the region of $[0.1,0.4]$, where the elastic-plastic transition occurs, is seen in the GRU similiarity scores, while the variance of the initial conditions is seen in the NODE similarity scores.

\begin{figure}[htb!]
\centering
\includegraphics[width=0.34\textwidth]{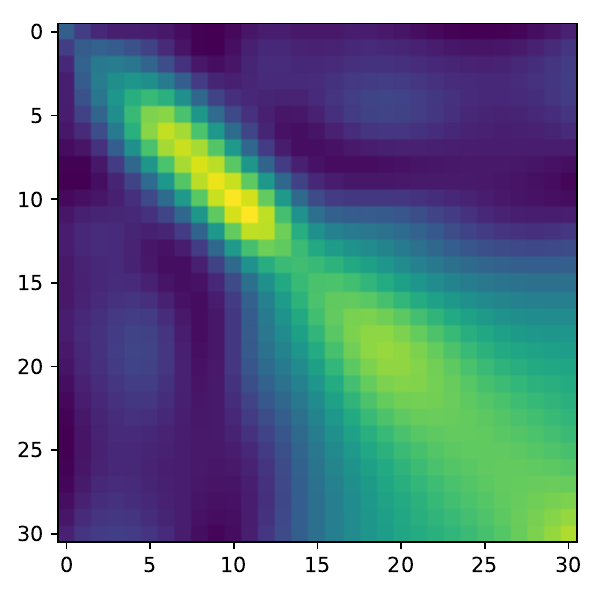}
\includegraphics[width=0.45\textwidth]{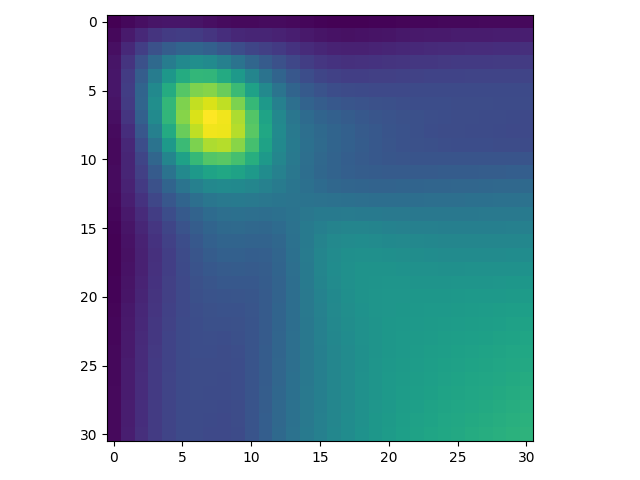}\\
\includegraphics[width=0.34\textwidth]{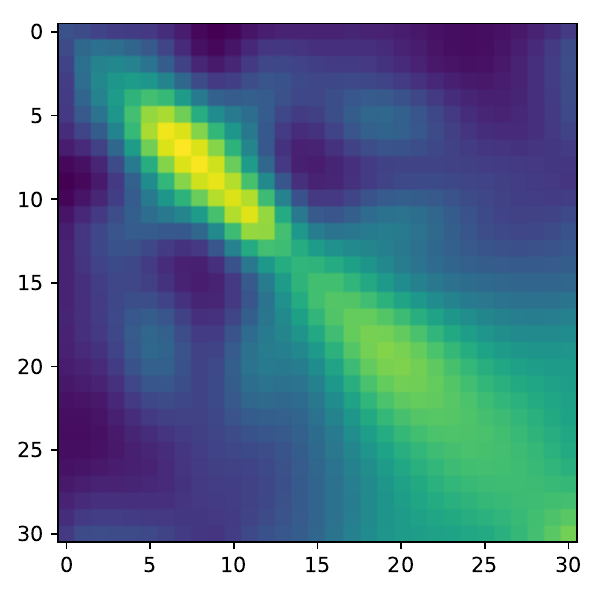}
\includegraphics[width=0.45\textwidth]{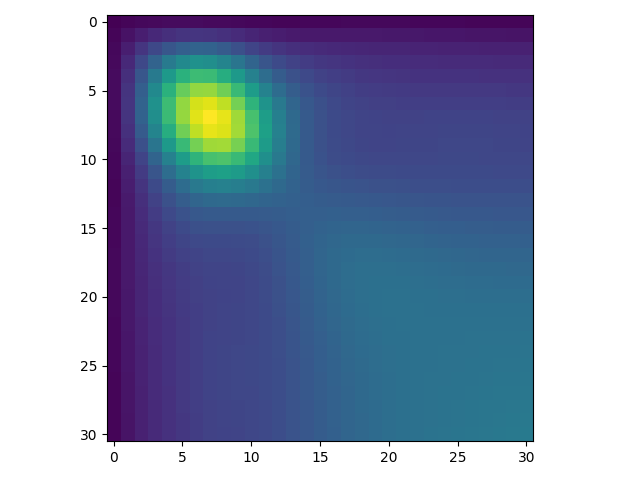}

\caption{GRU: comparison of posterior push-forward correlations over time, comparing HMC (left) versus SVGD (right) for $N_D = \frac{1}{2} N_w$ (top) and $N_D=2 N_w$ (bottom). The $x$- and $y$-axis values are time-step $n$.
}
\label{fig:rnn_covar}
\end{figure}

\begin{figure}[htb!]
\centering
\includegraphics[width=0.35\textwidth]{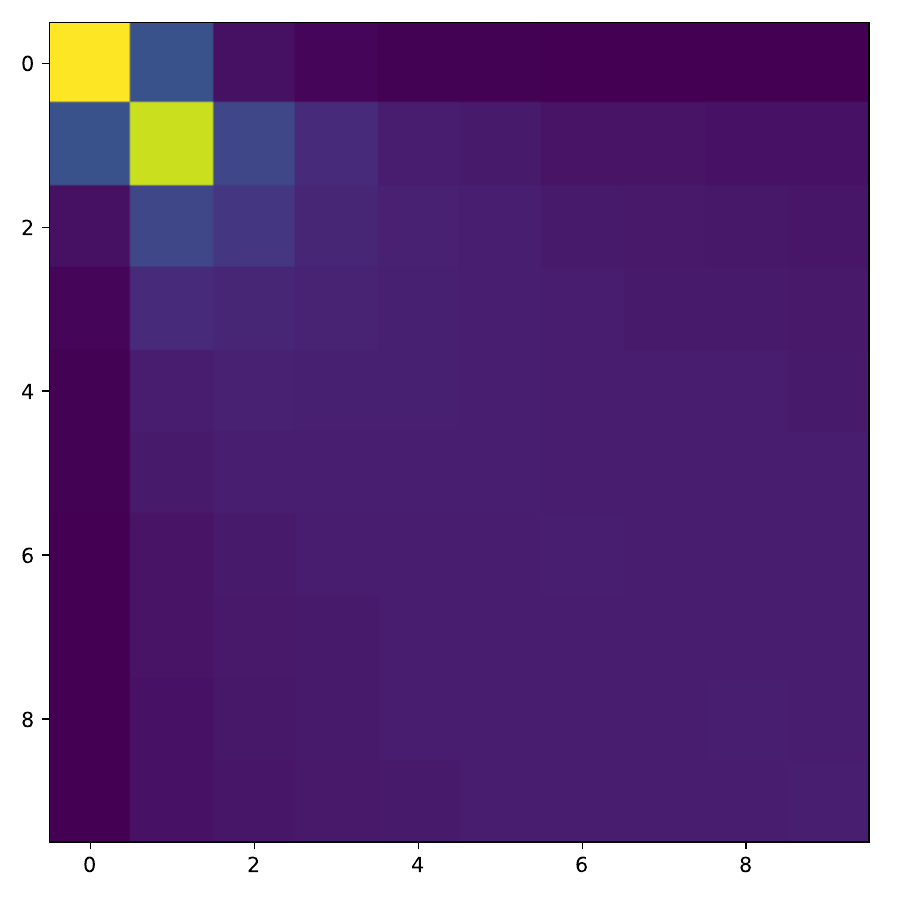}
\includegraphics[width=0.35\textwidth,trim={2.15cm 0 2.15cm 0},clip]{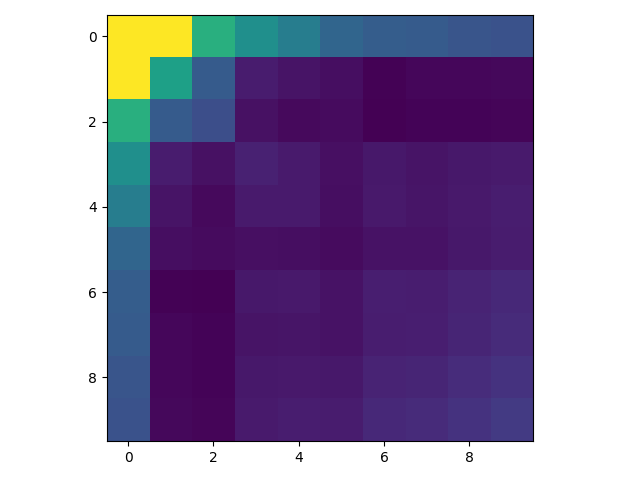}\\
\includegraphics[width=0.35\textwidth]{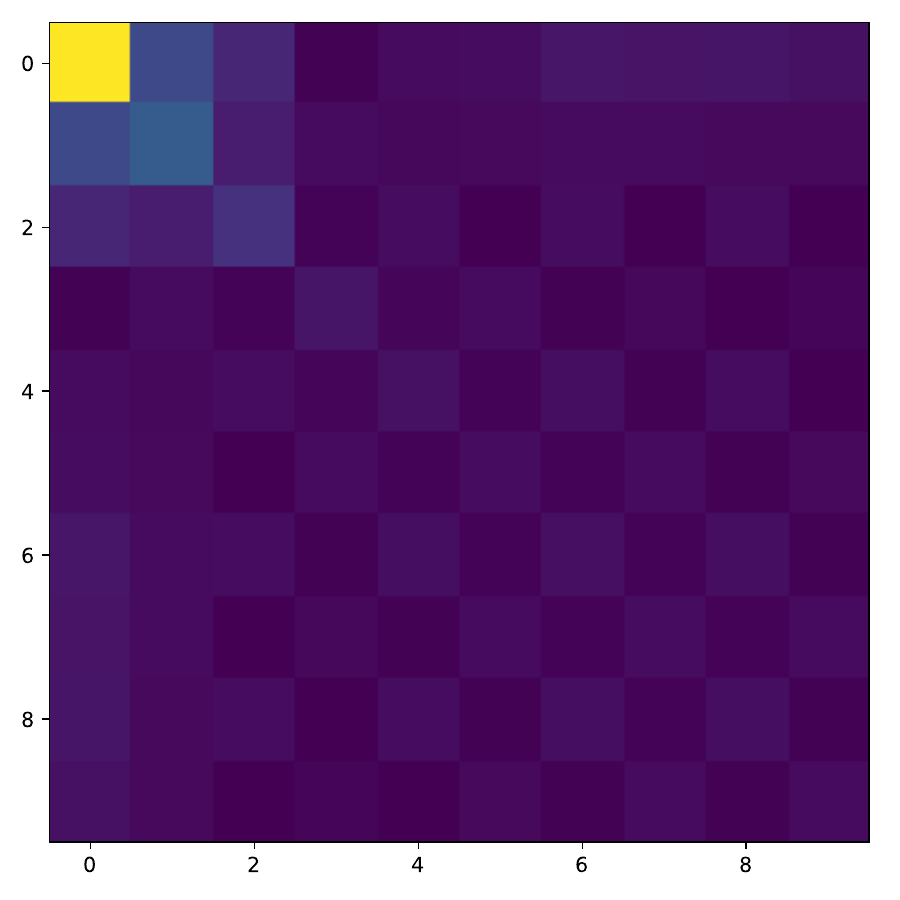}
\includegraphics[width=0.35\textwidth,trim={2.1cm 0 2.1cm 0},clip]{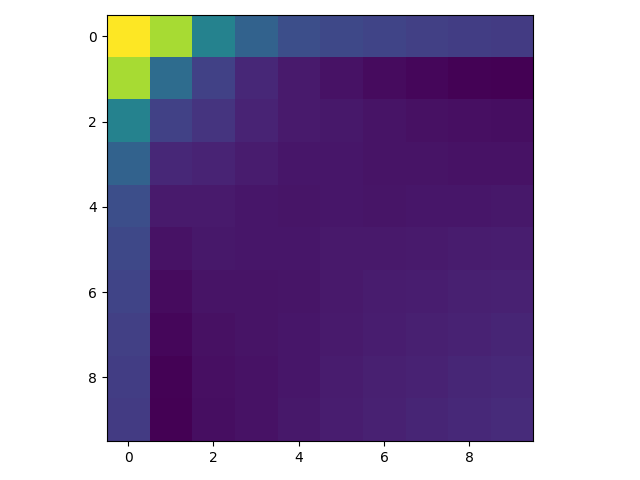}
\caption{NODE: comparison of posterior push-forward correlations over time, comparing HMC (left) versus SVGD (right) for $N_D = \frac{1}{2} N_w$ (top) and $N_D=2 N_w$ (bottom). The $x$- and $y$-axis values are time-step $n$.
}
\label{fig:node_covar}
\end{figure}

Finally, \fref{fig:rnn_covar} and \fref{fig:node_covar} show the average correlation of the predicted QoI at different
time steps, defined as
\begin{equation}
c_{mn} = \langle
[ \Delta \hat{\outputvector}(\inputvector_i, \timevector_i; \weight_{(j)})]_m
[ \Delta \hat{\outputvector}(\inputvector_i,\timevector_i;\weight_{(j)})]_n
\rangle
\quad \text{with} \quad
\Delta \hat{\outputvector}_i(\inputvector_i,\timevector_i;\weight_{(j)})  \equiv \hat{\outputvector}_i(\inputvector_i,\timevector_i;\weight_{(j)}) - \hat{\outputvector}(\inputvector_i,\timevector_i;\weight_\text{MAP})
\end{equation}
for two time steps $t_m$ and $t_n$, and where the average (inner product) is over weight/model realizations ${(j)}$ and input data samples $i$.
The plots show that HMC and SVGD predict similar covariance structures albeit with some qualitative differences.
For  GRU, the HMC covariance structure is more diagonally dominant, while SVGD appears to be capturing significant cross correlation in the elastic-plastic transition.
Likewise for the NODE, the HMC is diagonally dominant which SVGD appears to attribute more correlation of the initial conditions with all future states.

\clearpage % temporary
\subsection{Comparison of SVGD and pSVGD} \label{sec:comp_psvgd}

In this section we demonstration SVGD and pSVGD on a model infeasible for HMC due to the high number of parameters.
As mentioned in \sref{sec:gru}, we use a version of the GCNN-RNN crystal plasticity model operating on 3D data with 2609 parameters:  640 in convolution layers, 1680 in GRU, and rest in dense layers.
Due to memory constraints, we can only propagate 32 particles per MAP in SVGD on a single GPU.

\begin{figure}[htb!]
\centering
\includegraphics[width=0.48\textwidth]{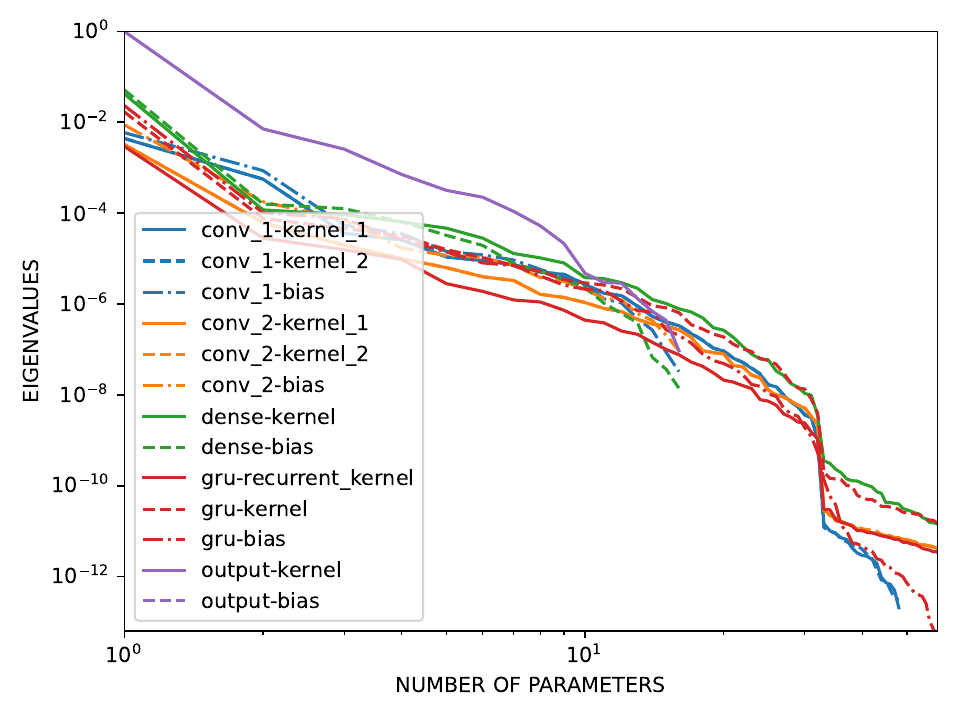}
\includegraphics[width=0.48\textwidth]{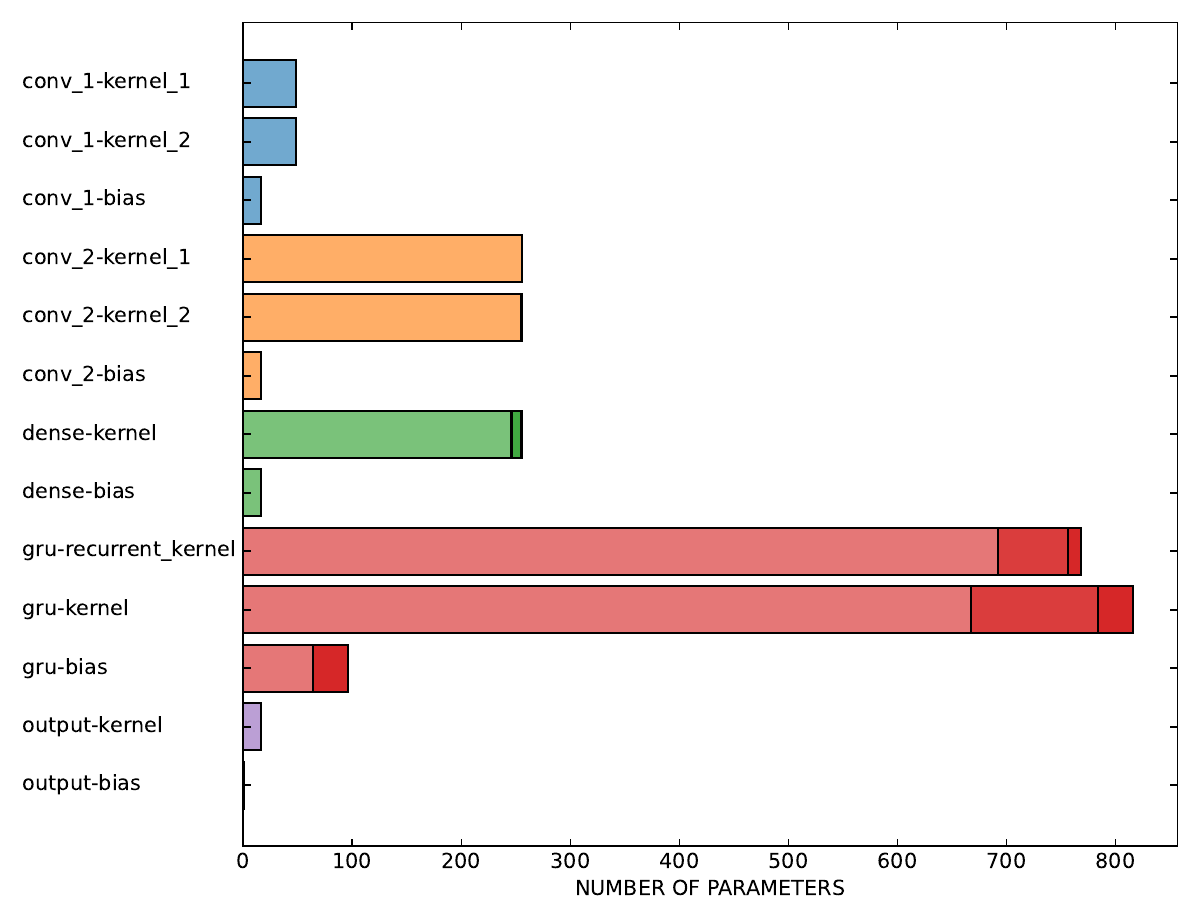}
\caption{3D GRU:  eigenspectrums (left) and energy per layer and  reduction profiles (right).
For the reduction profiles, the three models---full, 17\% reduction, and 32\% reduction---are shown by stacked bars.
} \label{fig:cp3d_energy}
\end{figure}

\fref{fig:cp3d_energy} shows the eigenspectrum of the Hessians of all the layers.
Note the large range and the steep decay in magnitude.
\fref{fig:cp3d_energy} also shows profile of reduction for two pSVGD cases: 3\% and 10\% reduction in the active weight space.
The parameters are primarily eliminated in the GRU which has the the largest proportion of the weight and also controls the stability of the time evolution.
Results at the high end of the eigenspectrum largely the same with $N_D = \{10, 100\}$ (not shown for brevity).

\begin{figure}[htb!]
\centering
\includegraphics[width=0.85\textwidth]{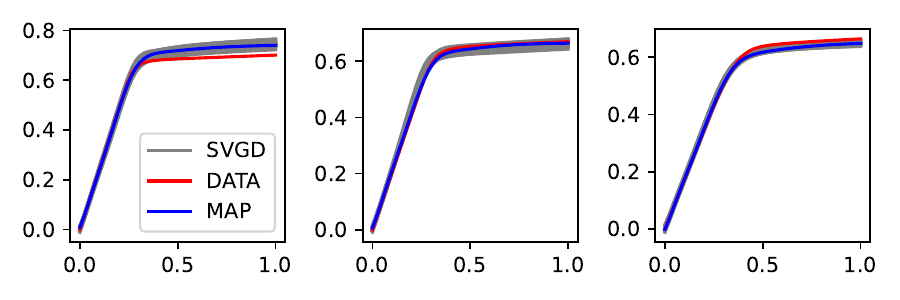}

\includegraphics[width=0.85\textwidth]{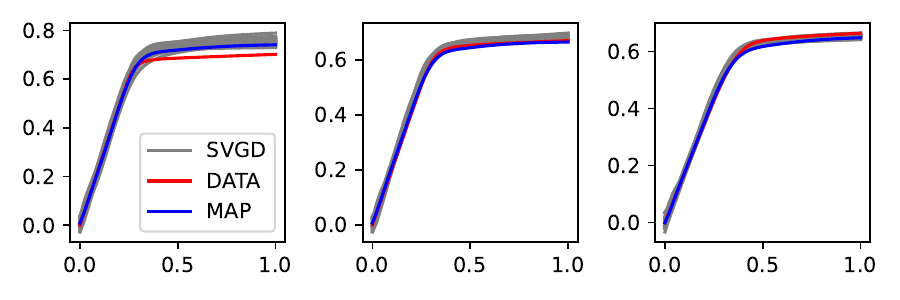}

\includegraphics[width=0.85\textwidth]{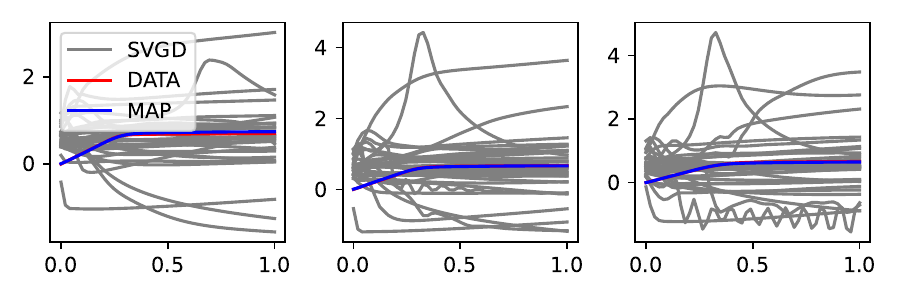}

\caption{3D GRU: posterior push-foward samples for three cases corresponding to minimum (left), median (center), and maximum (right) MAP discrepancy, for SVGD (top row) and pSVGD with 3\% (middle row) and 10\% reductions (bottom row).
} \label{fig:cp3d_pf}
\end{figure}

\fref{fig:cp3d_pf} shows the posterior push-forward for the three cases corresponding to minimum, median and maximum MAP discrepancy.
In this figure we show the bundle of posterior push-forward trajectories to illustrate the quality of these distributions for SVGD, and pSVGD with 3\% and 10\% reductions.
The 3\% reduction is visibly similar to the full SVGD; however, the realizations with 10\%  reduction indicate sampling of trajectories that appear to be marginally stable, which may be incurred from a relatively wide prior.

\begin{figure}[htb!]
\centering
\includegraphics[width=0.32\textwidth]{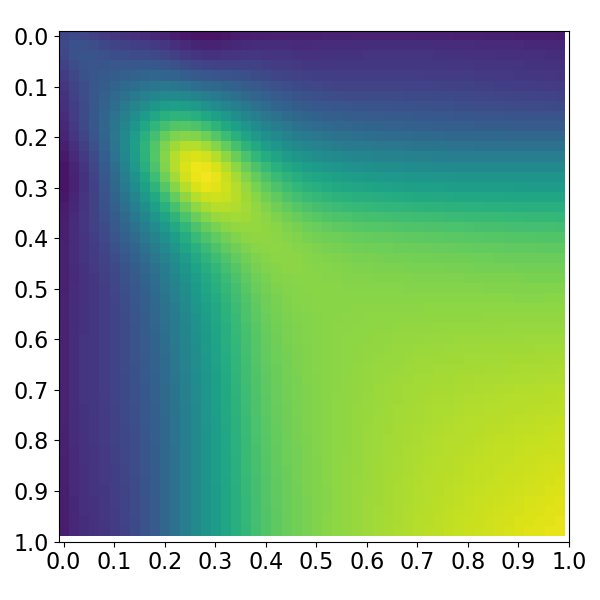}
\includegraphics[width=0.32\textwidth]{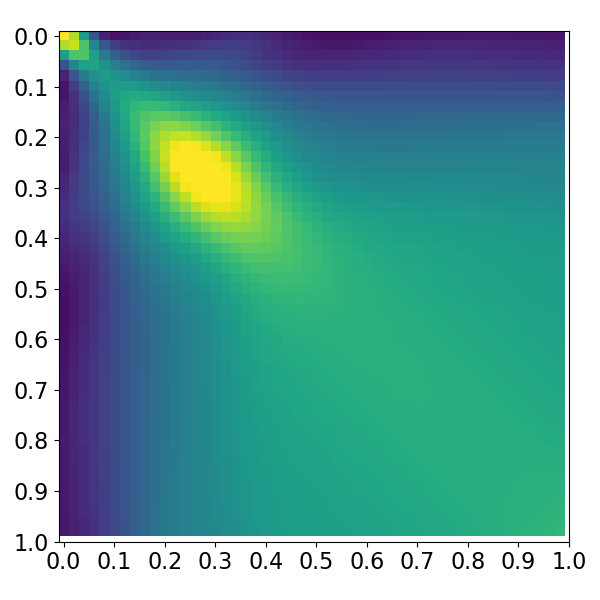}
\caption{3D GRU: comparison of posterior push-forward correlations over time, for SVGD (left) and pSVGD with 3\% reduction (right).
} \label{fig:cp3d_covar}
\end{figure}

\fref{fig:cp3d_covar} compares the posterior push-forward correlations of the SVGD and pSVGD with 3\% reduction.
The covariance structures are largely similar.
The pSVGD covariance structure show more variance at the initial state and less in the elastic-plastic transition regime.
In light of the benchmark illustration in \sref{sec:illustration} on the performance of pSVGD on a problem with multiple connected maxima of the posterior, and the indications of a qualitatively similar scenario for this exemplar shown in \fref{fig:rnn_submanifold}, it is not surprising that pSVGD gives a different PF covariance structure. At the same time, the pSVGD covariance structure is much closer to the HMC results in the 2D case in \fref{fig:rnn_covar}, suggesting the ability of pSVGD to achieve a more accurate posterior result due to its ability in achieving a higher particle-to-parameter ratio.

\begin{figure}[htb!]
\centering
\includegraphics[width=0.45\textwidth]{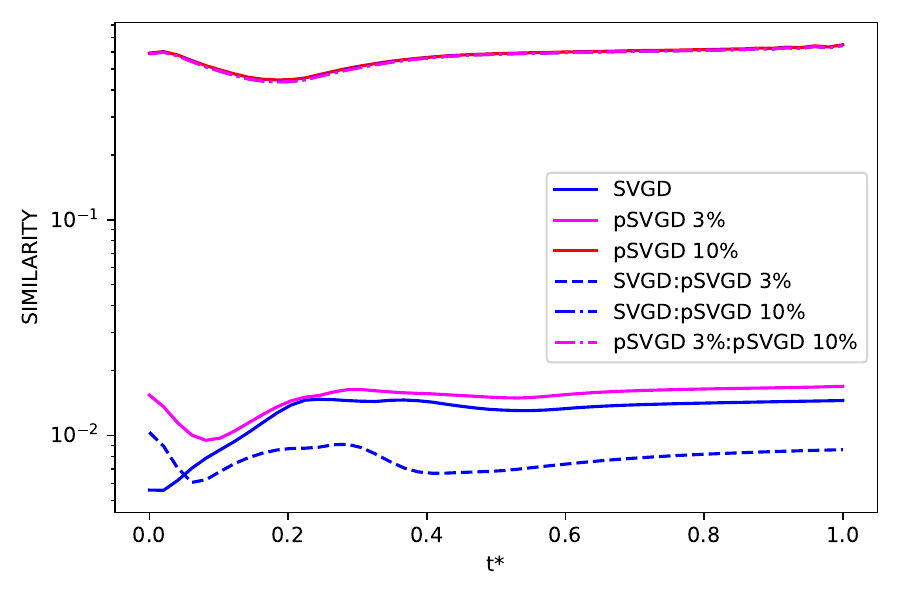}
\caption{3D GRU: KSS similarity measures between SVGD and data, and pSVGD and data (solid lines); and among these algorithms (dashed lines). A lower similarity score indicates a greater similarity.
} \label{fig:cp3d_dist}
\end{figure}

\fref{fig:cp3d_dist} shows the KSS similarity as defined earlier between the various posterior push-forward results compared to the data, and among results from the different algorithms.
For $N_D = 1000$, SVGD and pSVGD with 3\% reduction are nearly indistinquishable except near initial state, while both are far from the 10\% reduction.
\fref{fig:cp3d_dist} shows the effect of dataset size.
Clearly the posterior push-forward become farther from the data with less training data.
At the same time the difference between full and projected SVGD becomes less, as the prior becomes more influential.

\begin{figure}[htb!]
\centering
\includegraphics[width=0.45\textwidth]{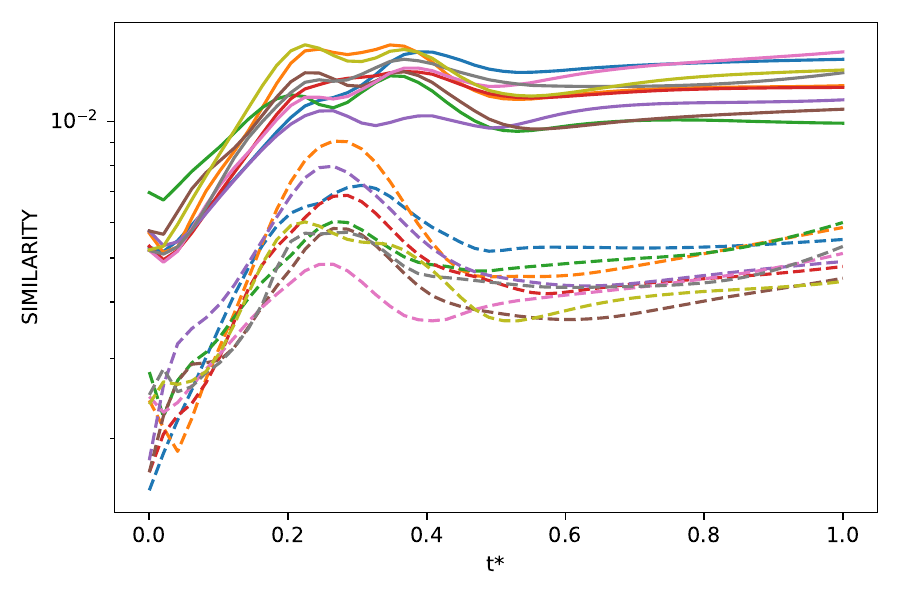}
\caption{3D GRU: KSS similarity measures for SVGD versus data based on an ensemble of different MAPs (solid) and between each SVGD result with the rest of the ensemble (dashed).
} \label{fig:cp3d_maps}
\end{figure}

To explore whether the limitation on the number of simultaneous particles could be ameliorated by sampling from more MAPs, we ran an ensemble of 9 independent MAPs.
\fref{fig:cp3d_maps} shows KSS similiarity across results from multiple  MAPs.
It appear that the MAPs produce essentially equivalent
results in terms of the posterior push-forward distributions.

\begin{figure}[htb!]
\centering
\includegraphics[width=0.45\textwidth]{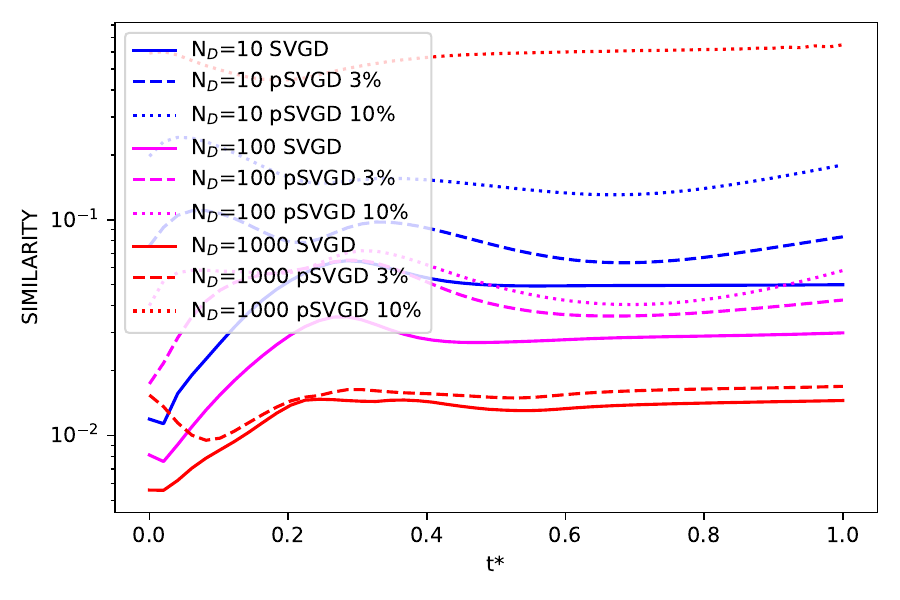}
\caption{3D GRU: KSS similarity measures between SVGD (solid), 3\% reduction pSVGD (dashed) and 10\% reduction pSVGD (dotted) versus data for $N_D = \{10,100,1000\}$ (blue, magenta, red). A lower similarity score indiciates a greater similarity.
} \label{fig:cp3d_ndata}
\end{figure}

\section{Conclusion} \label{sec:conclusion}
In this paper we employed two graph-based convolutional neural network (GCNN) architectures to model spatio-temporal process in two physical systems of practical interest.
The first model employed a gated recurrent unit (GRU) to simulate the stress response in polycrystals, while the second model was based on a neural ordinary differential equation (NODE) to predict diffusion and gas release in an intergranular network.
The discrepancy between the model and the data was framed in a Bayesian framework, which casts the neural network architecture to its Bayesian counterpart, and the resulting posterior parameter distribution was explored with both Hamiltonian Monte Carlo (HMC), Stein variational gradient descent (SVGD), and projected SVGD algorithms.
We showed that SVGD produced posterior push-forward distributions that are similar to the reference results obtained via HMC, for both the GCNN-GRU and the GCNN-NODE models.
Although the HMC chains were not well mixed due to the high-dimensionality of the models and the complexity of the posterior landscape and the SGVD particle ensembles were at best on par with the size of the parameter space, the posterior push-forward distribution profiles through time were similar with each other and consistent with observed data.
HMC results illustrated the convoluted seemingly low dimensional manifold of the likely model weights. We attribute this to both the slow mixing of the chains as well as to the highly irregular posterior landscape to be expected for NN models.

Our particle ensembles were limited by memory on a single GPU, which could be expanded by distributing the ensemble across multiple GPUs as suggested by Chen and Ghattas~\cite{chen2020projected}.
Being limited to less particles than dimensions, we may not have sampled the full covariance structure but the results did appear to produce posterior push-forward distributions that are consistent with the observed data.
The projection from pSVGD proved less effective than we hoped, in part due to impairing the stability of the time-evolution in the recurrent neural network model by pruning its weight space, but also due to the convoluted posterior manifold.
As illustrated in \sref{sec:illustration}, the convoluted submanifold likely does not yield to accurate reduction as a subspace.
Updating the projection as the particles move through the posterior landscape may help but the coordination of the particles constrains the pSVGD method to a single subspace for all particles, and is essentially a Laplace-like approximation.

In future work we will explore nonlinear model reduction techniques to further improve SVGD for these type of problems.

\section*{Acknowledgements}
Tensorflow \cite{tensorflow} and Spektral \cite{spektral} were used to implement the models and methods of this work.
This open-source software is gratefully acknowledged.
CS and REJ would like to thank Prof. M. Tonks (Univ. Florida) and his group for the inspiration for the gas release exemplar.
XH and JH acknowledge funding support  by the U.S. Department of Energy, Office of Science, Office of Advanced Scientific Computing Research, under Award Number DE-SC0021397.
This material is based upon work supported by the U.S. Department of Energy, Office of Science, Advanced Scientific Computing Research program.
Sandia National Laboratories is a multimission laboratory managed and operated by National Technology and Engineering Solutions of Sandia, LLC., a wholly owned subsidiary of Honeywell International, Inc., for the U.S. Department of Energy's National Nuclear Security Administration under contract DE-NA-0003525. This paper describes objective technical results and analysis. Any subjective views or opinions that might be expressed in the paper do not necessarily represent the views of the U.S. Department of Energy or the United States Government.

\end{document}